\documentclass{article}
\usepackage{arxiv}
\usepackage[utf8]{inputenc}
\usepackage[T1]{fontenc}
\usepackage{hyperref}
\usepackage{url}
\usepackage{booktabs}
\usepackage{amsfonts}
\usepackage{nicefrac}
\usepackage{microtype}
\usepackage[numbers]{natbib}
\usepackage{hyphenat}
\usepackage{doi}
\usepackage{float}
\usepackage{xcolor}
\usepackage{amsmath}
\usepackage{graphicx}
\usepackage{array}
\usepackage{tabularx}
\usepackage{makecell}
\usepackage{multirow}
\usepackage{longtable}
\usepackage{pdflscape}
\usepackage{fvextra}

\title{CBT-Audio: Evaluating Audio Language Models for Patient-Side Distress Intensity Estimation in CBT Session Recordings}

\author{
\begin{minipage}{0.95\textwidth}
\centering
Qixuan Hu~$^{1}$, Shuchang Ye~$^{1}$, Xumou Zhang~$^{1}$, Anastasia Serafimovska~$^{2}$,\\
Anastasia Suraev~$^{2}$, Amit Saha~$^{1}$, Ping-hsiu Lin~$^{4}$, Sydney Su,\\
Usman Naseem~$^{3}$, Adam G. Dunn~$^{5}$, Jinman Kim~$^{1}$
\end{minipage}
\\[1em]
\begin{minipage}{0.95\textwidth}
\centering
\small
$^{1}$School of Computer Science, Faculty of Engineering, University of Sydney, Australia\\
$^{2}$School of Psychology, Faculty of Science, University of Sydney, Australia\\
$^{3}$School of Computing, 
Faculty of Science and Engineering, Macquarie University, Australia\\
$^{4}$CHeBA (Centre for Healthy Brain Ageing), School of Clinical Medicine, Discipline of Psychiatry \& Mental Health, The University of New South Wales, Australia\\
$^{5}$Sydney School of Public Health, Faculty of Medicine and Health, University of Sydney, Australia
\texttt{qihu6986@uni.sydney.edu.au}
\end{minipage}
}

\date{}

\renewcommand{\shorttitle}{\textit{arXiv} Template}

\hypersetup{
pdftitle={CBT-Audio Archive},
pdfauthor={Qixuan Hu, Shuchang Ye, Xumou Zhang, Anastasia Serafimovska, Anastasia Suraev, Amit Saha, Ping-hsiu Lin, Sydney Su, Usman Naseem, Adam G. Dunn, Jinman Kim},
pdfkeywords={Audio Language Model, Cognitive Behavioural Therapy}
}

\begin{document}
\pagestyle{plain}  
\maketitle

\begin{abstract}
Cognitive behavioural therapy is widely used to help patients understand and manage psychological distress. It is often delivered through spoken conversation, where therapists attend not only to what patients say, but also to how they say it, because these cues can help therapists decide how to respond and adapt treatment. Progress in building AI systems for CBT remains largely limited to text, partly because most available datasets are text based and shareable spoken CBT data are scarce under ethical and privacy constraints. This creates a blind spot because text based models and evaluations cannot capture the mismatch between the transcript and the patient's voice, even though therapists often rely on this mismatch to understand patient distress. We introduce CBT-Audio, a dataset for evaluating patient distress estimation from spoken CBT sessions with audio language models. CBT-Audio contains 1,802 patient turns from 96 publicly available CBT recordings, with turn-level distress labels validated on an experts-annotated subset. We evaluate 10 open source audio language models under three input conditions, where models receive only patient audio, only the transcript, or both audio and transcript. Our results show that audio can provide useful information beyond text, especially when combined with transcripts. Adding audio to transcript input improves distress estimation over using the transcript alone in 8 of 10 model families, with significant gains in 4, and case studies show the clearest benefit when verbal content and vocal delivery diverge. CBT-Audio makes spoken patient behaviour measurable for AI evaluation in CBT-related tasks and supports future work on audio language models for mental health interaction.
\end{abstract}

\section{Introduction}
Cognitive Behavioral Therapy (CBT) is among the most widely practiced and empirically supported psychotherapies. It has shown efficacy for depression, anxiety, and a range of related disorders \citep{hofmann2012efficacy, butler2006empirical}. CBT is delivered through spoken conversation between therapist and patient. During sessions, the therapist tracks how the patient's emotional state shifts and adjusts the intervention accordingly \citep{westra2012therapist}. With the global shortage of trained mental health professionals \citep{ballout2025trauma, saraceno2007barriers} and recent work exploring Large Language Models (LLMs) for CBT assistance, response generation, and evaluation \citep{zhang2025cbt, hodson2024can, shen2024large}, AI systems are increasingly being studied as tools to support CBT practice, supervision, and quality evaluation.

Recent work has started applying LLMs to CBT-related tasks. CBT-Bench \citep{zhang2025cbt} introduced a structured benchmark spanning CBT based knowledge classification task, and therapeutic response generation, finding that LLMs handle knowledge questions well but struggle with complex therapeutic reasoning and response generation. On the data side, several efforts have constructed synthetic therapy dialogue datasets to address the scarcity of real counseling data. Cactus \citep{lee2024cactus} generated multi-turn CBT conversations using client personas and structured counseling strategies. DiaCBT \citep{zhou2025diacbt} extended this to multi-session dialogues guided by cognitive conceptualisation diagrams. Thousand Voices of Trauma \citep{bn2025thousand} produced synthetic conversations modeling emotional progression across trauma therapy stages. RealCBT \citep{wang2025feel} compared synthetic CBT sessions with CBT case studies or role play and found that synthetic dialogues, while fluent and coherent, tend to lack the emotional variability and reactivity patterns observed in therapy case studies done by real human experts, particularly on the patient side. Across all of these efforts, the input to the language models are limited to the text transcripts of therapy conversations.

Treating CBT sessions as text removes an entire modality of clinical information. Therapists are trained to attend to the mismatch between what a patient says and how they say it \citep{philippot2003nonverbal, foley2010nonverbal, gorawara2017exploring}. In a live session, the same words can carry different clinical meanings depending on how they are spoken. A patient saying ``I'm fine'' in a flat, withdrawn voice signals something different from the same words spoken calmly and steadily. A long pause before answering a difficult question can indicate avoidance, but this signal is often lost or simplified when sessions are modeled as text transcript. Rapid speech and rising pitch may reveal distress that the words themselves do not express. Vocal cues such as tone, pace, hesitation, tremor, and shifts in speech rate convey information about patient distress beyond what words alone provide \citep{cannizzaro2004voice, foley2010nonverbal, rocco2018beyond, kappen2024acoustic}. When models are evaluated only on transcripts, they cannot access these signals. Without an evaluation that includes spoken CBT session recordings, there is no way to measure whether models can use the vocal cues that therapists rely on.

\begin{figure}[!ht]
  \centering
  \includegraphics[width=0.99\linewidth, keepaspectratio]{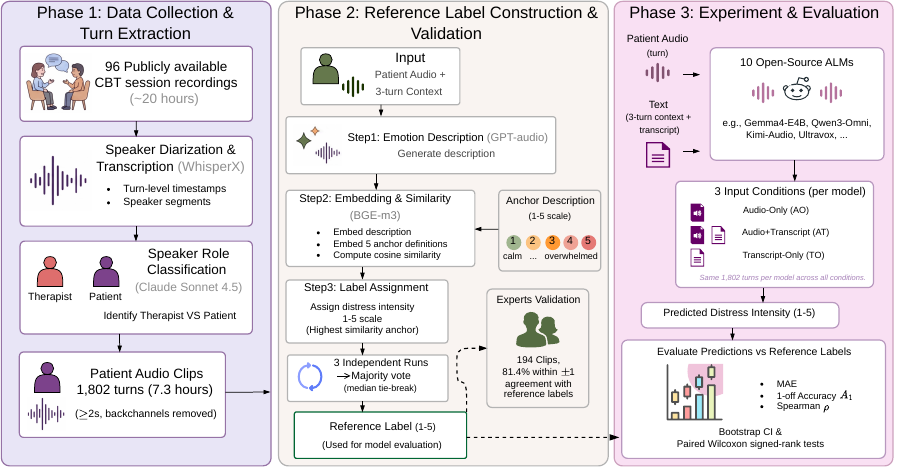}
  \caption{Pipeline from publicly available CBT sessions to model evaluation: (1) Patient turn extraction using speaker diarization, (2) GPT-audio based distress intensity labeling with expert validation, (3) Evaluation of ten ALMs across three input conditions.}
  \label{fig:pipeline}
\end{figure}

We introduce CBT-Audio, a dataset for testing whether audio language models (ALMs) can estimate patient distress intensity from spoken CBT session recordings. Figure~\ref{fig:pipeline} gives an overview of the dataset construction and evaluation pipeline. Our goal is not to build a clinical diagnostic system. Instead, we study a controlled evaluation question: when the same patient utterance is represented as audio, transcript, or both, do current open-source ALMs use the audio to improve ordinal distress estimation? This directly tests whether models benefit from access to the speech modality that transcript-only evaluations remove. CBT-Audio contains 1,802 patient-side utterances from 96 publicly available CBT educational recordings. Each utterance is labeled on a 1--5 distress intensity scale adapted from the Positive and Negative Affect Schedule (PANAS), a widely used measure for rating the intensity of positive and negative emotions, including distress \citep{watson1988development, crawford2004positive}. PANAS has also been used in prior CBT outcome and CBT-AI evaluation work \citep{lee2024cactus, zhou2025diacbt, fulmer2018using}. The labels are validated using ratings from five independent expert annotators. 

Our contributions are as follows.
\begin{enumerate}
  \item We introduce CBT-Audio, a dataset for evaluating patient distress estimation from spoken CBT sessions with audio language models. CBT-Audio provides source URLs, clip timestamps, transcripts, turn level distress labels and evaluation configuration, enabling reproducible evaluation.

  \item We show that current open source ALMs can use audio cues from CBT speech to estimate patient distress. Audio only input does not consistently outperform transcript only input, but adding audio to the transcript improves distress estimation in 8 of 10 model families, with significant gains in four. This shows that audio can carry cues indicating patient distress that complement the transcript, even when audio is not the stronger single modality.

  \item We provide a systematic evaluation of 10 open source ALMs under three input conditions, where models receive only patient audio, only the transcript, or both audio and transcript. Because each model is evaluated on the same patient turns across input conditions, this design allows us to compare the effect of input modality while keeping the task and data fixed.
\end{enumerate}

\section{Related Work}
\subsection{AI for CBT and Mental Health}

Recent work has applied LLMs to a growing range of CBT-related tasks. Several studies have evaluated how well current models can perform CBT itself. CBT-Bench \citep{zhang2025cbt} tested LLMs across CBT knowledge, cognitive distortion classification, and therapeutic response generation, finding that models fall short of human therapists when generating therapeutic responses but perform comparably on cognitive understanding tasks. Other studies reached the same picture where LLMs should not be relied on to lead CBT delivery, yet can offer reasonable suggestions for isolated cognitive tasks such as reframing unhelpful thoughts \citep{hodson2024can}, and paired with a CBT knowledge base they can produce plausible therapeutic dialogue \citep{shen2024large}. Several studies have built synthetic CBT dialogue datasets to train text-based conversational agents, since real therapy transcripts are scarce and raise ethical concerns. Cactus \citep{lee2024cactus} and DiaCBT \citep{zhou2025diacbt} generate multi-turn CBT conversations using client personas and cognitive conceptualization diagrams, and SMILE \citep{qiu2024smile} converts single-turn counseling exchanges into 55k multi-turn textual dialogue used to train a downstream mental health chatbot. Thousand Voices of Trauma \citep{bn2025thousand} takes a different angle, producing synthetic Prolonged Exposure dialogue transcripts for PTSD and releasing an accompanying benchmark for patient distress trajectories, though its labels are assigned at the session level over synthetic text rather than at the patient utterance level in real recordings. The same group behind Thousand Voices of Trauma later compared real and synthetic PE conversations and found that while surface linguistic features are replicated well, subtle therapeutic dynamics such as distress monitoring are not \citep{bn2025real}. RealCBT \citep{wang2025feel} reached a similar conclusion showing that synthetic CBT dialogues lack the emotional variability and reactivity patterns of real therapy sessions. Across these efforts, the input condition remains text. None incorporate audio from therapy sessions.

Outside of CBT specifically, speech-based approaches have been widely explored for mental health assessment. DAIC-WOZ \citep{gratch2014distress} is a commonly used corpus of clinical interview recordings that has supported work on depression detection from voice. A recent systematic review of 14 speech emotion recognition studies in mental health found that existing work mainly focuses on diagnostic classification, while tracking emotion changes over time remains an open direction \citep{jordan2025speech}. To our knowledge, no prior work studies turn-level distress changes within a single CBT session. 

Together, these areas leave an evaluation gap. Text-based CBT work does not assess vocal cues, while speech-based mental health work has focused mainly on diagnostic classification. Our work addresses this gap by measuring whether ALMs can estimate turn-level distress intensity in CBT from audio, transcripts, or both.

\subsection{Audio Language Models}
ALMs provide a natural testbed for this question because they can process spoken input and produce predictions. ALMs have progressed rapidly from early systems that bolted a speech encoder onto a text LLM to recent models that train audio representations at a scale comparable to text and vision. SALMONN \citep{tang2023salmonn} is an early example of the basic recipe, connecting a frozen Whisper encoder together with a BEATs encoder, one for linguistic content and one for paralinguistic cues, to a text LLM backbone. Subsequent work extended this pattern into unified multimodal models that handle audio alongside vision and video. MiniCPM-o \citep{yao2024minicpm}, Phi-4-Multimodal \citep{abouelenin2025phi}, and Qwen2.5-Omni \citep{xu2025qwen25omnitechnicalreport} all follow the same encoder--adapter--LLM recipe but differ in how much they adapt the underlying language model, ranging from LoRA-only modality adapters over a frozen backbone to full unfreezing across multimodal training data. In parallel, lighter-weight audio-focused models such as Ultravox \citep{fixie2024ultravox}, Voxtral \citep{liu2025voxtral}, and Kimi-Audio \citep{ding2025kimi} apply the same recipe with minimal projectors over frozen or partially-updated LLM backbones. More recent work moves in the opposite direction, training audio encoders at much larger scale: Audio Flamingo 3 \citep{goel2025audio} trains its own encoder on top of Whisper, and Qwen3-Omni \citep{xu2025qwen3} replaces Whisper entirely with a custom audio encoder trained from scratch on 20 million hours of audio. Gemma-4 \citep{deepmind2026gemma4e4b} extends this trend into the on-device multimodal setting. Across all of these models, the audio pathway is projected into the embedding space of a text LLM and trained end-to-end with it on tasks such as automatic speech recognition, speech translation, and spoken question answering. Only 3 of 10 documented the inclusion of emotion recognition data during training and 5 of 10 include emotion recognition from speech in their evaluation task. The corpora used for both training and evaluation, such as IEMOCAP \citep{busso2008iemocap}, MELD \citep{poria2019meld}, and RAVDESS \citep{livingstone2018ryerson}, provide categorical emotion labels over single short utterances drawn from general-domain speech, and none target therapeutic or clinical conversation.

\section{Dataset Construction}

The gaps identified above point to a need for evaluating current open-source ALMs on within-session distress intensity estimation in CBT. This section describes how we construct such a dataset from publicly available CBT session recordings, covering turn extraction, reference labeling, and expert validation.

\paragraph{Source material.} The dataset is built from 96 publicly available CBT session recordings hosted on YouTube, totalling approximately 20 hours of audio. The recordings cover CBT role-play demonstrations and case-study walkthroughs produced as educational content. All recordings are in English. No sessions involve real patients or contain personally identifiable information. We release source URLs, clip timestamps, reference labels and related code so that users can reconstruct the dataset locally. The complete list of 96 source recordings is provided in Appendix~\ref{app:source-recordings}.

\paragraph{Turn extraction.} Each recording is processed into speaker turns using WhisperX \citep{bain2023whisperx} with Whisper-large-v3 \citep{radford2023robust} for transcription and forced alignment, and pyannote speaker-diarization-3.1 \citep{plaquet2023powerset, bredin2023pyannote} for diarization. WhisperX produces word-level timestamps, and pyannote assigns each word to an anonymous speaker label. We then merge consecutive same-speaker segments separated by less than five seconds into turns. To identify which speaker is the therapist and which is the patient, we pass the transcripts to Claude Sonnet 4.5, which classifies each speaker based on conversational content and turn-taking patterns. We retain all patient turns whose audio is at least two seconds long, filtering out single-word backchannels such as "yeah" or "okay." The resulting dataset contains 1,802 patient turns across the 96 sessions, totalling 7.3 hours of patient audio with a mean clip duration of 14.6 seconds and a median of 9.8 seconds.

\subsection{Reference Label Construction}

Each patient turn receives a reference label between 1-5 for perceived patient distress intensity. The five anchor definitions are adapted from the affect-intensity descriptors of the PANAS framework \citep{watson1988development, crawford2004positive} and are shown in Table~\ref{tab:anchors}. We construct reference labels using the semantic similarity rating (SSR) approach of \citet{maier2025llms}. For each patient turn, we prompt OpenAI's GPT-audio 1.5 model with the audio clip and the three most recent turns of conversational context. The model is asked to produce a clear description of the patient's emotional state, using vocal evidence such as tone, pace, and voice quality. The exact prompts are in Appendix~\ref{app:ssr-description-prompt}. We encode the description with the BGE-m3 sentence embedding model \citep{bge-m3} and compute cosine similarity to embeddings of the five anchor definitions. The label is the anchor with the highest similarity. We repeat the full pipeline three times independently and take the per-turn majority vote, using the median as a tiebreaker since the scale is ordinal. Across the 1{,}802 turns, the three runs agree within $\pm 1$ on 94.3\% of turns, and produce similar label distributions (Appendix~\ref{app:run-distribution}). SSR produces a variable label distribution across the five levels (Appendix~\ref{app:ssr-vs-numeric}).

\subsection{Expert Annotation and Validation}
Distress intensity in therapy is inherently subjective, so we validate whether the reference labels align with experts judgment. A subset of 194 patient clips across 8 sessions was annotated by five experts with clinical, counseling, and mental health research backgrounds. The panel included expertise in clinical psychology, neuro-psychotherapy, counseling training, psychiatry research, and health psychology research support. Experts received the same inputs available to models, including the audio clip, transcript, and conversational context. They rated each clip on the same 1--5 scale and anchor definitions without seeing reference labels or other experts' ratings. On the expert-labeled subset, the five-expert consensus agrees with our reference labels within $\pm 1$ on 81.4\% of clips. Inter-rater agreement across experts is moderate, with Krippendorff's $\alpha = 0.439$ for ordinal data and a mean pairwise quadratic weighted $\kappa = 0.441$. The model ranking from the expert-labeled subset is also highly consistent with the ranking from the full dataset (Spearman $\rho = 0.882$, $p < 0.001$). Together, these results support using the reference labels as a reliable basis for this evaluation. Additional human validation results are reported in Appendix~\ref{app:human-validation}.

\begin{table}[!htbp]
  \centering
  \caption{Distress-intensity anchor descriptions used for SSR label construction, model evaluation prompts, and expert annotation. Adapted from the affect-intensity descriptors of the PANAS framework.}
  \label{tab:anchors}
  \small
    \begin{tabular}{c p{0.78\linewidth}}
    \toprule
    \textbf{Level} & \textbf{Anchor description} \\
    \midrule
    1 & Calm, at ease, with no emotional distress \\
    2 & Slightly uneasy, with mild emotional activation \\
    3 & Moderately distressed, with clear emotional activation \\
    4 & Significantly distressed, struggling to maintain composure \\
    5 & Overwhelmed with intense emotional distress. \\
    \bottomrule
  \end{tabular}
\end{table}

\section{Experiments}

\subsection{Setup}

\paragraph{Task.} Given a patient utterance from a CBT session and the three most recent preceding turns as conversational context, a model must output a single integer from 1 to 5 estimating the patient's distress intensity. The context is transcript-only and includes both therapist and patient turns. Models receive the full 1--5 scale with anchor definitions (Table~\ref{tab:anchors}) and are asked to respond with a single number. The exact prompts are in Appendix~\ref{app:evaluation-prompts}. Unlike label construction, where SSR separates description from scale assignment to avoid a single model's scale-use patterns propagating into all reference labels, evaluation includes the scale directly. This is because each model's use of the scale is part of what we are measuring, and routing all model outputs through a shared embedding model would partly benchmark that embedding model rather than the ALMs themselves.

\paragraph{Input conditions.} Each ALM is evaluated under three input conditions. In the audio-only condition (AO), the model receives the patient audio clip and the three-turn conversational context, but not the reference transcript of the current patient turn. The model may still infer lexical content from the audio through its own audio encoder. We do not prevent this, because that is part of what an audio language model is designed to do. In the audio+transcript (AT) condition, the model receives the audio clip, the context, and the dialogue transcript of the current patient turn. In the transcript-only (TO) condition, the model receives the context and the dialogue transcript of the current patient turn, but no audio. All three conditions share the same system prompt and scale definitions. Running all three conditions on the same model and the same patient utterances means that difference in performance comes from the input condition, not the model backbone.

\paragraph{Models.} We evaluate ten open-source ALMs: SALMONN-7B \citep{tang2023salmonn}, MiniCPM-o 2.6 \citep{yao2024minicpm}, Phi-4-Multimodal \citep{abouelenin2025phi}, Qwen2.5-Omni-7B \citep{xu2025qwen25omnitechnicalreport}, Qwen3-Omni-30B-A3B \citep{xu2025qwen3}, Kimi-Audio-7B \citep{ding2025kimi}, Ultravox-v0.5-Llama-3.1-8B \citep{fixie2024ultravox}, Audio Flamingo 3 \citep{goel2025audio}, Voxtral-Mini \citep{liu2025voxtral}, and Gemma-4-E4B \citep{deepmind2026gemma4e4b}. We selected these models because they are publicly available, support audio-in text-out inference, and represent recent major directions in open ALM development. The set includes early general-audio LLMs, compact multimodal models, end-to-end omni models, speech-understanding models, and recent audio foundation models. It also covers different audio integration strategies, including Whisper-style encoders, BEATs-style audio encoders, conformer-based encoders, and modality-specific LoRA. This allows us to test whether patient distress estimation is supported across current ALM designs, rather than within a single architecture family. We focus on open-source models because CBT recordings contain emotionally sensitive speech. Sending such recordings to commercial inference APIs raises privacy and data-retention concerns that are difficult to resolve for downstream users in clinical settings \citep{bucher2025s, voultsiou2026systematic, obradovich2024opportunities, hua2025scoping, asman2025responsible}. Open-source models that can be run locally or self-hosted offer a practical path toward privacy-preserving evaluation and deployment. Table~\ref{tab:results} lists the models evaluated alongside their results, and Appendix~\ref{app:model-details} provides more model-level details on architecture.

\paragraph{Metrics and statistical testing.} Our primary metric is Mean Absolute Error (MAE) on the 1--5 scale, which measures prediction error directly in scale units and is computable per turn, supporting paired statistical tests. We additionally report Spearman rank correlation $\rho$, Quadratic Weighted Cohen's $\kappa$ (QWK), Root Mean Squared Error (RMSE), and 1-off accuracy $A_1$, defined as the fraction of predictions that fall within one level of the reference label on the 1--5 scale. For each metric we report 95\% percentile bootstrap confidence intervals over turns ($B = 1{,}000$ resamples). Pairwise comparisons between input conditions of the same model use paired Wilcoxon signed-rank tests on per-turn absolute errors with a significance threshold of $p < 0.05$.

\subsection{Results and Discussion}

\paragraph{Overall model performance.} Table~\ref{tab:results} reports MAE, $A_1$, and Spearman $\rho$ for all ten audio models under each input condition. Qwen3-Omni-30B-A3B and Gemma-4-E4B are the two strongest models, with audio-only MAE of 0.785 and 0.824 respectively. These are also the only two mixture-of-experts models in our evaluation and the most recently released. Ultravox-v0.5 (0.912) and Qwen2.5-Omni-7B (0.963) form a second cluster. The remaining models fall between 1.06 and 1.16, with SALMONN-7B as a clear outlier at 1.606. Among the dense models, parameter count does not clearly correlate with performance. Voxtral-Mini at 3.6 billion parameters performs comparably to models twice its size, and six models that share a similar 7 to 8 billion parameter scale span a wide performance range, suggesting that architectural and training choices matter more than scale at this size. The best single configuration across all models and conditions is Gemma-4-E4B with audio+transcript input (MAE 0.720, $A_1$ 90.0\%). The contrast between Qwen3-Omni and Gemma-4-E4B suggests that the best audio-only model is not necessarily the best audio+transcript model. Qwen3-Omni performs best with audio alone, which may reflect its audio-focused design: its technical report describes a custom AuT audio encoder trained from scratch on 20 million hours of supervised audio, together with strong results across many audio and audio-visual benchmarks. Gemma-4-E4B is strongest when transcript information is available, suggesting that its advantage on this task may come less from audio side alone and more from using the transcript to map verbal content onto patient distress. In this setting, audio may still help by adding cues about how the utterance was delivered, while the transcript preserves what was said.

\begin{table}[!ht]
\centering
\caption{Results for ten ALMs under three input conditions. MAE (lower is better) is the primary metric; $A_1$ is the percentage of predictions within one level of the reference label (higher is better); $\rho$ is Spearman rank correlation (higher is better). Models are sorted by Audio+Transcript MAE. \textbf{Bold} indicates the best input condition per model for each metric. \underline{Underline} indicates the best model per condition for each metric. $^{*}$ marks model families where Audio+Transcript significantly improves over Transcript-only based on a paired Wilcoxon signed-rank test over per-turn absolute errors ($p < 0.05$). Full results and 95\% bootstrap CIs are in Appendix~\ref{app:full-results}.}
\label{tab:results}
{\footnotesize
\setlength{\tabcolsep}{2pt}
\begin{tabular}{l r r ccc ccc ccc}
\toprule
& & & \multicolumn{3}{c}{Audio-only} & \multicolumn{3}{c}{Transcript-only} & \multicolumn{3}{c}{Audio+Transcript} \\
\cmidrule(lr){4-6} \cmidrule(lr){7-9} \cmidrule(lr){10-12}
Model & Year & Params & MAE$\downarrow$ & $A_1$$\uparrow$ & $\rho$$\uparrow$ & MAE$\downarrow$ & $A_1$$\uparrow$ & $\rho$$\uparrow$ & MAE$\downarrow$ & $A_1$$\uparrow$ & $\rho$$\uparrow$ \\
\midrule
Gemma-4-E4B$^{*}$         & 2026 & $\sim$4.5B & 0.824 & 86.5 & 0.483 & \underline{0.746} & \underline{89.0} & \underline{0.578} & \underline{\textbf{0.720}} & \underline{\textbf{90.0}} & \underline{\textbf{0.587}} \\
Qwen3-Omni$^{*}$         & 2025 & 30B (3B) & \underline{\textbf{0.785}} & \underline{\textbf{89.9}} & \underline{\textbf{0.583}} & 0.873 & 84.2 & 0.531 & 0.813 & 88.2 & 0.583 \\
Ultravox-v0.5       & 2025 & $\sim$8B & 0.912 & 80.0 & 0.419 & \textbf{0.883} & \textbf{82.0} & \textbf{0.524} & 0.895 & 81.4 & 0.513 \\
Qwen2.5-Omni        & 2025 & $\sim$7B & 0.963 & 80.5 & 0.431 & 0.941 & 79.8 & 0.463 & \textbf{0.931} & \textbf{80.6} & \textbf{0.464} \\
Kimi-Audio-7B$^{*}$      & 2025 & $\sim$7B & 1.062 & 78.1 & 0.261 & 1.133 & 69.6 & \textbf{0.387} & \textbf{0.987} & \textbf{81.4} & 0.339 \\
MiniCPM-o 2.6       & 2025 & $\sim$8B & 1.129 & 70.2 & 0.394 & 1.089 & 70.4 & \textbf{0.528} & \textbf{1.079} & \textbf{72.9} & 0.503 \\
Voxtral-Mini        & 2025 & $\sim$3B & 1.106 & \textbf{75.7} & 0.279 & 1.115 & 73.1 & 0.353 & \textbf{1.104} & 74.9 & \textbf{0.372} \\
Audio Flamingo 3$^{*}$    & 2025 & $\sim$8B & 1.156 & 69.8 & 0.303 & 1.143 & 71.3 & 0.328 & \textbf{1.106} & \textbf{71.8} & \textbf{0.410} \\
Phi-4-Multimodal    & 2025 & $\sim$5.6B & \textbf{1.114} & \textbf{71.1} & 0.278 & 1.156 & 65.9 & 0.357 & 1.137 & 68.8 & \textbf{0.359} \\
SALMONN-7B          & 2023 & $\sim$7B & 1.606 & 34.2 & 0.019 & \textbf{1.585} & \textbf{35.4} & \textbf{0.082} & 1.601 & 34.4 & 0.038 \\
\bottomrule
\end{tabular}
}
\end{table}
\begin{figure}[!htbp]
    \centering
    \includegraphics[width=0.8\linewidth]{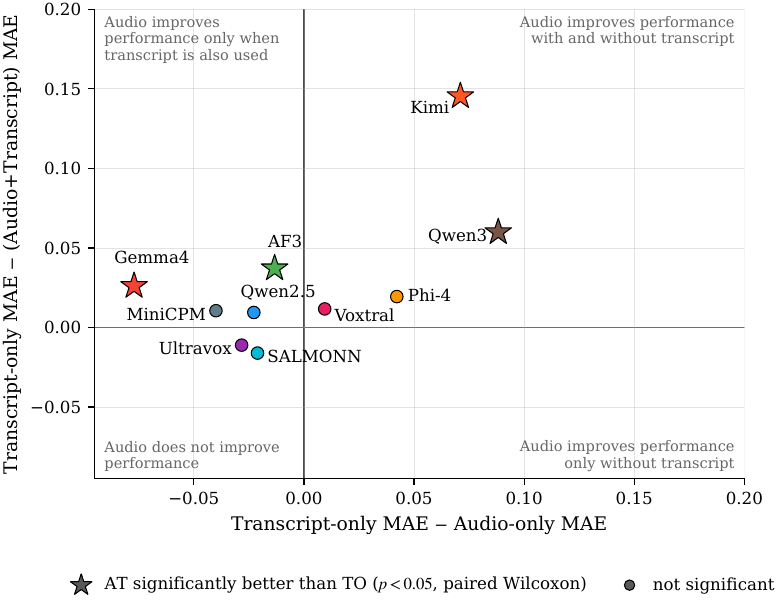}
    \caption{
    Modality improvement landscape. The x-axis shows how much audio-only improves MAE over transcript-only, and the y-axis shows how much audio+transcript improves MAE over transcript-only. Higher values indicate larger improvements over transcript-only. Star markers indicate significant audio+transcript improvement over transcript-only under a paired Wilcoxon signed-rank test on per-turn absolute errors ($p < 0.05$).
    }
    \label{fig:modality_gain_landscape}
\end{figure}

\paragraph{Audio combined with text beats either modality alone.}
Comparing audio-only with transcript-only within each model family, the results are mixed. Audio-only is significantly better in three families and significantly worse in three, with four showing no significant difference ($p < 0.05$, Wilcoxon signed-rank). This suggests that audio alone does not reliably outperform transcript input on this task. However, combining audio with transcript gives a clearer benefit. Audio+transcript improves over transcript-only in eight of ten model families, with significant gains in four: Audio Flamingo 3, Kimi-Audio-7B, Gemma-4-E4B, and Qwen3-Omni-30B-A3B. This overall pattern is shown in Figure~\ref{fig:modality_gain_landscape}, where most models lie above the horizontal zero line. Only Ultravox-v0.5 and SALMONN-7B have transcript-only as their best setting, and SALMONN-7B is the only model where audio+transcript is significantly worse than transcript-only.

Figure~\ref{fig:modality_gain_landscape} also shows that audio can add value in more than one way. For some models, audio is useful both as a standalone input and when combined with the transcript. For other models, audio alone is weaker than transcript-only, but still improves performance when added to the transcript. This suggests that audio does not need to be the stronger single modality to provide useful complementary information. In this setting, the transcript captures what was said, while the audio may add cues about how it was said. These results suggest that therapy audio carries information that text alone does not capture, and that many current ALMs can use this information when both modalities are available.

Our results show that audio provides valuable information for estimating patient distress, most clearly when both audio and transcript are given to the model. To better understand in what scenarios audio modality helps most and what information each modality contributes, we examine individual case studies below.

\begin{figure}[!ht]
  \centering
  \includegraphics[width=\linewidth,keepaspectratio]{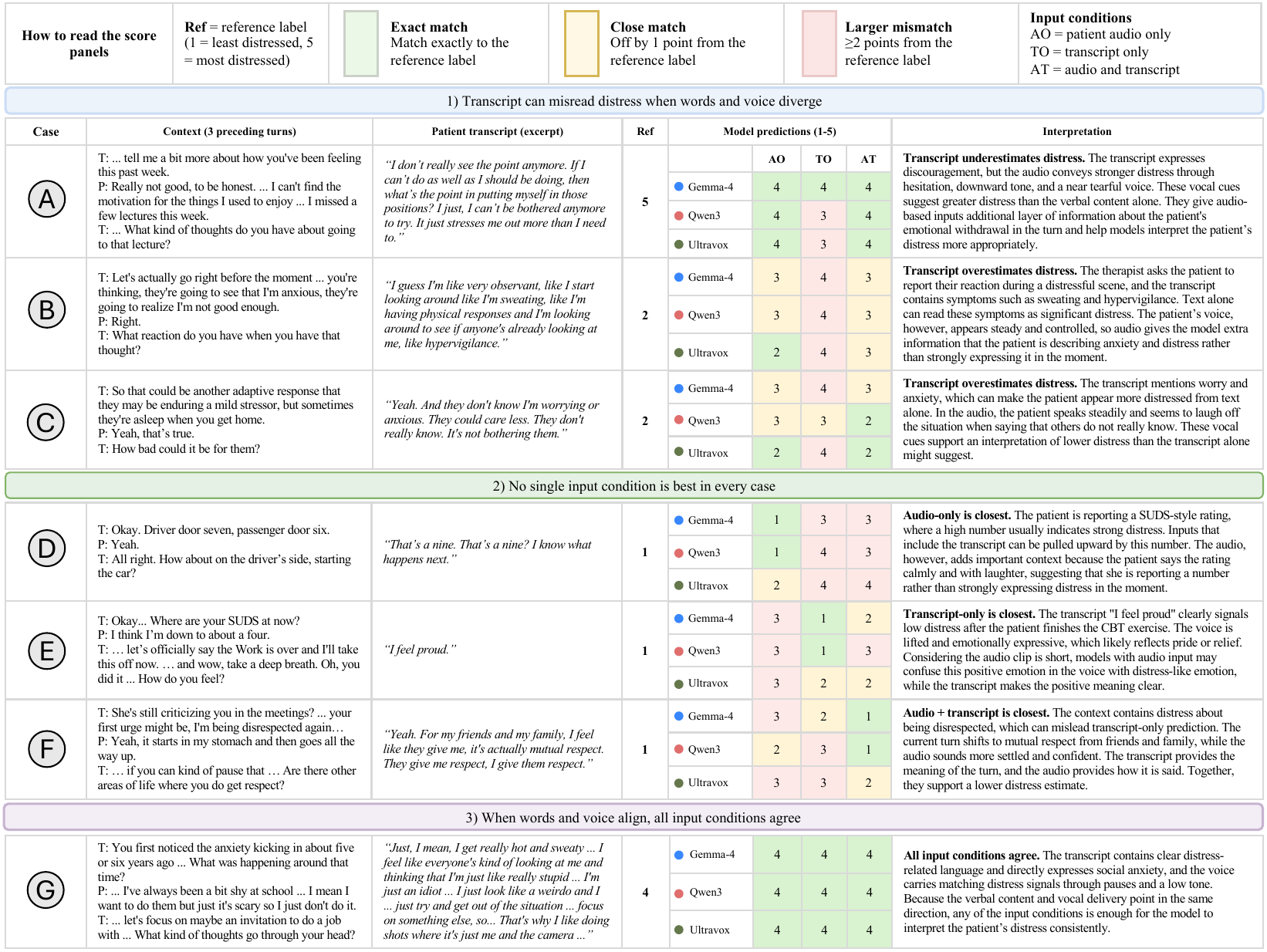}
  \caption{Case studies showing how input condition affects distress estimation. Each case shows the preceding conversational context, patient transcript, reference label, and predictions from three representative models under AO, TO, and AT input. Context and transcript are excerpted, full versions are in Appendix~\ref{app:full-case-studies}.}
  \label{fig:casestudy}
\end{figure}

\subsubsection{Case Studies}
Figure~\ref{fig:casestudy} shows seven case studies and compares predictions from the three top-performing models (Gemma-4-E4B, Qwen3-Omni, and Ultravox-v0.5). These case studies were reviewed with experts to check that the interpretation of vocal delivery were clinically plausible.

Cases A to D show where audio cues improve estimation relative to transcript-only input. In Case A, the patient says that she cannot find the motivation to keep trying. The transcript alone misses hesitation and a near-tearful voice, which leads transcript-only models to underestimate her distress. Cases B and C show the opposite pattern. The transcript contains words that can suggest stronger distress, such as sweating, worrying, and feeling anxious. However, the patient appears steadier in voice and pace. In Case C, the patient also appears to laugh off the situation, suggesting uneasiness but not the level of distress implied by the transcript-only prediction. Case D is a distinctive example where the transcript itself can be misleading and can cause models to give an inaccurate estimate. The patient is rating anxiety for imagined situations, and the answer nine would usually indicate high anxiety and distress. However, in this turn, the patient speaks with a steady tone and laughter. Audio-only models capture this lighter delivery and move closer to the reference label, while transcript-based models are pulled toward a higher distress estimate.

Cases E to G show that audio does not always improve the estimate as a single modality. Case E follows the end of a CBT exercise where the patient has worked through a difficult task. The patient says ``I feel proud'' with a lifted and confident voice, and does not appear to express distress in the turn. Transcript-only models correctly predict calm (1), while models with audio access, especially audio-only models, overestimate the distress level. Case F shows a complementary pattern. The patient describes mutual respect within her family in a steady, flat vocal tone. This may suggest a disconnect between the positive content and her vocal delivery, but not clear distress in the moment. Both audio-only and transcript-only predictions overestimate distress. Case G shows where the transcript alone contains enough emotional cues, and adding audio does not change the model's estimation. 

\paragraph{Why how the patient says it matters for distress estimation in CBT.} 
Taken together, these case studies show that audio contributes most when \textit{what} the patient says and \textit{how} they say it diverge. In Cases A through D, vocal delivery suggests a different distress level from the transcript alone, and audio moves the prediction closer to the reference label. In Cases E through G, delivery and content align more closely, and the model performs generally better when given both text and audio. In CBT, patient distress can appear in tone, hesitation, pace, and withdrawal, not only in the words themselves \citep{foley2010nonverbal, tonti2016rate, philippot2003nonverbal, gorawara2017exploring, voss2024mental}. Adding audio therefore helps the model estimate distress when verbal content and vocal delivery diverge. Our results and case studies suggest that current ALMs can capture some vocal information in patient speech, and that audio should be treated as an important modality alongside transcripts when training and evaluating AI models for CBT distress estimation.

\section{Limitations}
CBT-Audio has several limitations. First, all 96 source recordings are in English. Second, turn extraction relies on automatic diarization and transcription, which can introduce errors, especially for overlapping speech or quiet utterances. We filtered out turns shorter than two seconds to reduce the impact of diarization noise. Third, the source recordings are educational CBT content, including role-plays and case-study walkthroughs. The patient-side speaker is played by an actor, trainee or clinician rather than a person seeking treatment. Although these recordings reflect CBT interaction patterns, they may not capture the full emotional range of therapy in real world. Finally, our labels measure perceived distress from observable interaction, not clinical diagnosis or patient self-report. CBT-Audio should therefore be used to evaluate model sensitivity to expressed distress in spoken CBT interaction, not as a substitute for clinical assessment.

\section{Ethics and Data Availability}
CBT-Audio is built from publicly available CBT educational videos hosted on YouTube. The recordings are role-plays and case-study walkthroughs, and the patient is played by an actor, a trainee, or a clinician. The data does not include real patient identifiers. Throughout this paper, patient distress intensity refers to the level of distress expressed by the patient-side speaker in a given utterance. CBT-Audio is intended for research evaluation only, its perceived distress labels should not be treated as clinical ground truth or used to support individual-level clinical assessments. We are currently preparing the complete public release of the dataset metadata. We do not redistribute raw audio files or clips but release derived research metadata and reproducibility code, including source URLs, timestamps, turn-level transcripts, labels, and evaluation configuration. Transcripts are included only for reproducibility and controlled evaluation.


\section{Conclusion}

Most current AI work on CBT is shaped by text based data, even though CBT is a spoken interaction. This creates an important gap because patient distress is not always visible from words alone, and therapists often attend to the mismatch between what patients say and how they sound. CBT-Audio helps address this gap by making the mismatch between what patients say and how they say it measurable. Our findings show that audio is not simply a replacement for transcript input. Audio only input does not consistently outperform transcript only input, but audio can still add useful information when combined with the transcript. This suggests that current audio language models can use vocal cues in therapeutic speech as complementary signals for patient distress that are not fully captured by text. Future CBT models should therefore be designed and evaluated not only as text models, but as models of spoken therapeutic interaction. CBT-Audio provides a foundation for this direction. It can support future evaluation of audio language models, development of models that better use vocal cues, and research on safer audio aware systems for mental health interaction.

\bibliographystyle{plainnat}
\bibliography{main}

@article{watson1988development,
  title={Development and validation of brief measures of positive and negative affect: the PANAS scales.},
  author={Watson, David and Clark, Lee Anna and Tellegen, Auke},
  journal={Journal of personality and social psychology},
  volume={54},
  number={6},
  pages={1063},
  year={1988},
  publisher={American Psychological Association}
}

@article{crawford2004positive,
  title={The Positive and Negative Affect Schedule (PANAS): Construct validity, measurement properties and normative data in a large non-clinical sample},
  author={Crawford, John R and Henry, Julie D},
  journal={British journal of clinical psychology},
  volume={43},
  number={3},
  pages={245--265},
  year={2004},
  publisher={Wiley Online Library}
}

@article{ballout2025trauma,
  title={Trauma, mental health workforce shortages, and health equity: A crisis in public health},
  author={Ballout, Suha},
  journal={International Journal of Environmental Research and Public Health},
  volume={22},
  number={4},
  pages={620},
  year={2025},
  publisher={MDPI}
}

@article{westra2012therapist,
  title={Therapist emotional reactions and client resistance in cognitive behavioral therapy.},
  author={Westra, Henny A and Aviram, Adi and Connors, Laura and Kertes, Angela and Ahmed, Mariyam},
  journal={Psychotherapy},
  volume={49},
  number={2},
  pages={163},
  year={2012},
  publisher={Educational Publishing Foundation}
}

@article{saraceno2007barriers,
  title={Barriers to improvement of mental health services in low-income and middle-income countries},
  author={Saraceno, Benedetto and van Ommeren, Mark and Batniji, Rajaie and Cohen, Alex and Gureje, Oye and Mahoney, John and Sridhar, Devi and Underhill, Chris},
  journal={The Lancet},
  volume={370},
  number={9593},
  pages={1164--1174},
  year={2007},
  publisher={Elsevier}
}

@inproceedings{zhang2025cbt,
  title={CBT-bench: Evaluating large language models on assisting cognitive behavior therapy},
  author={Zhang, Mian and Yang, Xianjun and Zhang, Xinlu and Labrum, Travis and Chiu, Jamie C and Eack, Shaun M and Fang, Fei and Wang, William Yang and Chen, Zhiyu},
  booktitle={Proceedings of the 2025 Conference of the Nations of the Americas Chapter of the Association for Computational Linguistics: Human Language Technologies (Volume 1: Long Papers)},
  pages={3864--3900},
  year={2025}
}

@inproceedings{lee2024cactus,
  title={Cactus: Towards psychological counseling conversations using cognitive behavioral theory},
  author={Lee, Suyeon and Mac Kim, Sunghwan and Kim, Minju and Kang, Dongjin and Yang, Dongil and Kim, Harim and Kang, Minseok and Jung, Dayi and Kim, Min Hee and Lee, Seungbeen and others},
  booktitle={Findings of the Association for Computational Linguistics: EMNLP 2024},
  pages={14245--14274},
  year={2024}
}

@article{zhou2025diacbt,
  title={DiaCBT: A Long-Periodic Dialogue Corpus Guided by Cognitive Conceptualization Diagram for CBT-based Psychological Counseling},
  author={Zhou, Yougen and Zhou, Ningning and Chen, Qin and Zhou, Jie and Zhou, Aimin and He, Liang},
  journal={arXiv preprint arXiv:2509.02999},
  year={2025}
}

@inproceedings{bn2025thousand,
    title={Thousand Voices of Trauma: A Large-Scale Synthetic Dataset for Modeling Prolonged Exposure Therapy Conversations},
    author={Suhas BN and Andrew M. Sherrill and Rosa I. Arriaga and Christopher Wiese and Saeed Abdullah},
    booktitle={The Thirty-ninth Annual Conference on Neural Information Processing Systems Datasets and Benchmarks Track},
    year={2025},
    url={https://openreview.net/forum?id=qrFvHgZa7l}
}

@article{wang2025feel,
  title={Feel the difference? a comparative analysis of emotional arcs in real and LLM-generated CBT sessions},
  author={Wang, Xiaoyi and Zhang, Jiwei and Zhang, Guangtao and Guo, Honglei},
  journal={arXiv preprint arXiv:2508.20764},
  year={2025}
}

@article{hofmann2012efficacy,
  title={The efficacy of cognitive behavioral therapy: A review of meta-analyses},
  author={Hofmann, Stefan G and Asnaani, Anu and Vonk, Imke JJ and Sawyer, Alice T and Fang, Angela},
  journal={Cognitive therapy and research},
  volume={36},
  number={5},
  pages={427--440},
  year={2012},
  publisher={Springer}
}

@article{butler2006empirical,
  title={The empirical status of cognitive-behavioral therapy: A review of meta-analyses},
  author={Butler, Andrew C and Chapman, Jason E and Forman, Evan M and Beck, Aaron T},
  journal={Clinical psychology review},
  volume={26},
  number={1},
  pages={17--31},
  year={2006},
  publisher={Elsevier}
}

@article{cannizzaro2004voice,
  title={Voice acoustical measurement of the severity of major depression},
  author={Cannizzaro, Michael and Harel, Brian and Reilly, Nicole and Chappell, Phillip and Snyder, Peter J},
  journal={Brain and cognition},
  volume={56},
  number={1},
  pages={30--35},
  year={2004},
  publisher={Elsevier}
}

@article{kappen2024acoustic,
  title={Acoustic and prosodic speech features reflect physiological stress but not isolated negative affect: a multi-paradigm study on psychosocial stressors},
  author={Kappen, Mitchel and Vanhollebeke, Gert and Van Der Donckt, Jonas and Van Hoecke, Sofie and Vanderhasselt, Marie-Anne},
  journal={Scientific Reports},
  volume={14},
  number={1},
  pages={5515},
  year={2024},
  publisher={Nature Publishing Group UK London}
}

@book{philippot2003nonverbal,
  title={Nonverbal behavior in clinical settings},
  author={Philippot, Pierre and Feldman, Robert S and Coats, Erik J},
  year={2003},
  publisher={Oxford University Press}
}

@article{foley2010nonverbal,
  title={Nonverbal communication in psychotherapy},
  author={Foley, Gretchen N and Gentile, Julie P},
  journal={Psychiatry (Edgmont)},
  volume={7},
  number={6},
  pages={38},
  year={2010}
}

@article{gorawara2017exploring,
  title={Exploring physicians’ verbal and nonverbal responses to cues/concerns: Learning from incongruent communication},
  author={Gorawara-Bhat, Rita and Hafskjold, Linda and Gulbrandsen, Paul and Eide, Hilde},
  journal={Patient education and counseling},
  volume={100},
  number={11},
  pages={1979--1989},
  year={2017},
  publisher={Elsevier}
}

@article{bucher2025s,
  title={“It’s Not Only Attention We Need”: Systematic Review of Large Language Models in Mental Health Care},
  author={Bucher, Andreas and Egger, Sarah and Vashkite, Inna and Wu, Wenyuan and Schwabe, Gerhard},
  journal={JMIR mental health},
  volume={12},
  pages={e78410},
  year={2025},
  publisher={JMIR Publications Toronto, Canada}
}

@article{voultsiou2026systematic,
  title={A Systematic Review of Large Language Models in Mental Health: Opportunities, Challenges, and Future Directions},
  author={Voultsiou, Evdokia and Moussiades, Lefteris},
  journal={Electronics},
  volume={15},
  number={3},
  pages={524},
  year={2026},
  publisher={MDPI}
}

@article{obradovich2024opportunities,
  title={Opportunities and risks of large language models in psychiatry},
  author={Obradovich, Nick and Khalsa, Sahib S and Khan, Waqas U and Suh, Jina and Perlis, Roy H and Ajilore, Olusola and Paulus, Martin P},
  journal={NPP—Digital Psychiatry and Neuroscience},
  volume={2},
  number={1},
  pages={8},
  year={2024},
  publisher={Springer International Publishing Cham}
}

@article{hua2025scoping,
  title={A scoping review of large language models for generative tasks in mental health care},
  author={Hua, Yining and Na, Hongbin and Li, Zehan and Liu, Fenglin and Fang, Xiao and Clifton, David and Torous, John},
  journal={npj Digital Medicine},
  volume={8},
  number={1},
  pages={230},
  year={2025},
  publisher={Nature Publishing Group UK London}
}

@article{asman2025responsible,
  title={Responsible design, integration, and use of generative AI in mental health},
  author={Asman, Oren and Torous, John and Tal, Amir},
  journal={JMIR Mental Health},
  volume={12},
  number={1},
  pages={e70439},
  year={2025},
  publisher={JMIR Publications Inc., Toronto, Canada}
}

@article{hodson2024can,
  title={Can large language models replace therapists? Evaluating performance at simple cognitive behavioral therapy tasks},
  author={Hodson, Nathan and Williamson, Simon},
  journal={JMIR AI},
  volume={3},
  number={1},
  pages={e52500},
  year={2024},
  publisher={JMIR Publications Inc., Toronto, Canada}
}

@inproceedings{shen2024large,
  title={Are large language models possible to conduct cognitive behavioral therapy?},
  author={Shen, Hao and Li, Zihan and Yang, Minqiang and Ni, Minghui and Tao, Yongfeng and Yu, Zhengyang and Zheng, Weihao and Xu, Chen and Hu, Bin},
  booktitle={2024 IEEE International Conference on Bioinformatics and Biomedicine (BIBM)},
  pages={3695--3700},
  year={2024},
  organization={IEEE}
}

@inproceedings{qiu2024smile,
  title={Smile: Single-turn to multi-turn inclusive language expansion via chatgpt for mental health support},
  author={Qiu, Huachuan and He, Hongliang and Zhang, Shuai and Li, Anqi and Lan, Zhenzhong},
  booktitle={Findings of the Association for Computational Linguistics: EMNLP 2024},
  pages={615--636},
  year={2024}
}

@article{bn2025real,
  title={How real are synthetic therapy conversations? evaluating fidelity in prolonged exposure dialogues},
  author={Bn, Suhas and Mattioli, Dominik and Sherrill, Andrew M and Arriaga, Rosa I and Wiese, Christopher and Abdullah, Saeed},
  journal={Findings of the Association for Computational Linguistics: EMNLP},
  volume={2025},
  pages={20986--20995},
  year={2025}
}

@inproceedings{gratch2014distress,
  title={The distress analysis interview corpus of human and computer interviews.},
  author={Gratch, Jonathan and Artstein, Ron and Lucas, Gale M and Stratou, Giota and Scherer, Stefan and Nazarian, Angela and Wood, Rachel and Boberg, Jill and DeVault, David and Marsella, Stacy and others},
  booktitle={Lrec},
  volume={14},
  pages={3123--3128},
  year={2014},
  organization={Reykjavik}
}

@article{jordan2025speech,
  title={Speech emotion recognition in mental health: Systematic review of voice-based applications},
  author={Jordan, Eric and Terrisse, Rapha{\"e}l and Lucarini, Valeria and Alrahabi, Motasem and Krebs, Marie-Odile and Descl{\'e}s, Julien and Lemey, Christophe},
  journal={JMIR mental health},
  volume={12},
  number={1},
  pages={e74260},
  year={2025},
  publisher={JMIR Publications Inc., Toronto, Canada}
}

@article{tang2023salmonn,
  title={Salmonn: Towards generic hearing abilities for large language models},
  author={Tang, Changli and Yu, Wenyi and Sun, Guangzhi and Chen, Xianzhao and Tan, Tian and Li, Wei and Lu, Lu and Ma, Zejun and Zhang, Chao},
  journal={arXiv preprint arXiv:2310.13289},
  year={2023}
}

@article{yao2024minicpm,
  title={MiniCPM-V: A GPT-4V Level MLLM on Your Phone},
  author={Yao, Yuan and Yu, Tianyu and Zhang, Ao and Wang, Chongyi and Cui, Junbo and Zhu, Hongji and Cai, Tianchi and Li, Haoyu and Zhao, Weilin and He, Zhihui and others},
  journal={arXiv preprint arXiv:2408.01800},
  year={2024}
}

@article{abouelenin2025phi,
  title={Phi-4-mini technical report: Compact yet powerful multimodal language models via mixture-of-loras},
  author={Abouelenin, Abdelrahman and Ashfaq, Atabak and Atkinson, Adam and Awadalla, Hany and Bach, Nguyen and Bao, Jianmin and Benhaim, Alon and Cai, Martin and Chaudhary, Vishrav and Chen, Congcong and others},
  journal={arXiv preprint arXiv:2503.01743},
  year={2025}
}

@misc{xu2025qwen25omnitechnicalreport,
      title={Qwen2.5-Omni Technical Report}, 
      author={Jin Xu and Zhifang Guo and Jinzheng He and Hangrui Hu and Ting He and Shuai Bai and Keqin Chen and Jialin Wang and Yang Fan and Kai Dang and Bin Zhang and Xiong Wang and Yunfei Chu and Junyang Lin},
      year={2025},
      eprint={2503.20215},
      archivePrefix={arXiv},
      primaryClass={cs.CL},
      url={https://arxiv.org/abs/2503.20215},
}

@article{liu2025voxtral,
  title={Voxtral},
  author={Liu, Alexander H and Ehrenberg, Andy and Lo, Andy and Denoix, Cl{\'e}ment and Barreau, Corentin and Lample, Guillaume and Delignon, Jean-Malo and Chandu, Khyathi Raghavi and von Platen, Patrick and Muddireddy, Pavankumar Reddy and others},
  journal={arXiv preprint arXiv:2507.13264},
  year={2025}
}

@misc{fixie2024ultravox,
  title={Ultravox: A fast multimodal llm for real-time voice},
  author={Fixie, AI},
  year={2024},
  howpublished = {\url{https://huggingface.co/fixie-ai/ultravox-v0_5-llama-3_1-8b}},
  note = {Official project website: \url{https://ultravox.ai}. Model evaluated: ultravox-v0\_5-llama-3\_1-8b}
}

@article{ding2025kimi,
  title={Kimi-audio technical report},
  author={Ding, Ding and Ju, Zeqian and Leng, Yichong and Liu, Songxiang and Liu, Tong and Shang, Zeyu and Shen, Kai and Song, Wei and Tan, Xu and Tang, Heyi and others},
  journal={arXiv preprint arXiv:2504.18425},
  year={2025}
}

@article{goel2025audio,
  title={Audio flamingo 3: Advancing audio intelligence with fully open large audio language models},
  author={Goel, Arushi and Ghosh, Sreyan and Kim, Jaehyeon and Kumar, Sonal and Kong, Zhifeng and Lee, Sang-gil and Yang, Chao-Han Huck and Duraiswami, Ramani and Manocha, Dinesh and Valle, Rafael and others},
  journal={arXiv preprint arXiv:2507.08128},
  year={2025}
}

@article{xu2025qwen3,
  title={Qwen3-omni technical report},
  author={Xu, Jin and Guo, Zhifang and Hu, Hangrui and Chu, Yunfei and Wang, Xiong and He, Jinzheng and Wang, Yuxuan and Shi, Xian and He, Ting and Zhu, Xinfa and others},
  journal={arXiv preprint arXiv:2509.17765},
  year={2025}
}

@misc{deepmind2026gemma4e4b,
  author = {{Google DeepMind}},
  title = {Gemma 4 E4B Instruct (gemma-4-E4B-it)},
  year = {2026},
  publisher = {Hugging Face},
  howpublished = {\url{https://huggingface.co/google/gemma-4-E4B-it}},
  note = {Official release announcement on Hugging Face: \url{https://huggingface.co/blog/gemma4}. Model evaluated: gemma-4-E4B-it}
}

@article{busso2008iemocap,
  title={IEMOCAP: Interactive emotional dyadic motion capture database},
  author={Busso, Carlos and Bulut, Murtaza and Lee, Chi-Chun and Kazemzadeh, Abe and Mower, Emily and Kim, Samuel and Chang, Jeannette N and Lee, Sungbok and Narayanan, Shrikanth S},
  journal={Language resources and evaluation},
  volume={42},
  number={4},
  pages={335--359},
  year={2008},
  publisher={Springer}
}

@inproceedings{poria2019meld,
  title={Meld: A multimodal multi-party dataset for emotion recognition in conversations},
  author={Poria, Soujanya and Hazarika, Devamanyu and Majumder, Navonil and Naik, Gautam and Cambria, Erik and Mihalcea, Rada},
  booktitle={Proceedings of the 57th annual meeting of the association for computational linguistics},
  pages={527--536},
  year={2019}
}

@article{livingstone2018ryerson,
  title={The Ryerson Audio-Visual Database of Emotional Speech and Song (RAVDESS): A dynamic, multimodal set of facial and vocal expressions in North American English},
  author={Livingstone, Steven R and Russo, Frank A},
  journal={PloS one},
  volume={13},
  number={5},
  pages={e0196391},
  year={2018},
  publisher={Public Library of Science}
}

@article{maier2025llms,
  title={LLMs Reproduce Human Purchase Intent via Semantic Similarity Elicitation of Likert Ratings},
  author={Maier, Benjamin F and Aslak, Ulf and Fiaschi, Luca and Rismal, Nina and Fletcher, Kemble and Luhmann, Christian C and Dow, Robbie and Pappas, Kli and Wiecki, Thomas V},
  journal={arXiv preprint arXiv:2510.08338},
  year={2025}
}

@article{bain2023whisperx,
  title={WhisperX: Time-Accurate Speech Transcription of Long-Form Audio},
  author={Bain, Max and Huh, Jaesung and Han, Tengda and Zisserman, Andrew},
  journal={INTERSPEECH 2023},
  year={2023},
  publisher={ISCA}
}

@inproceedings{radford2023robust,
  title={Robust speech recognition via large-scale weak supervision},
  author={Radford, Alec and Kim, Jong Wook and Xu, Tao and Brockman, Greg and McLeavey, Christine and Sutskever, Ilya},
  booktitle={International conference on machine learning},
  pages={28492--28518},
  year={2023},
  organization={PMLR}
}

@inproceedings{plaquet2023powerset,
  title={Powerset multi-class cross entropy loss for neural speaker diarization},
  author={Plaquet, Alexis and Bredin, Herv{\'e}},
  booktitle={24th Interspeech Conference (INTERSPEECH 2023)},
  pages={3222--3226},
  year={2023},
  organization={ISCA}
}

@inproceedings{bredin2023pyannote,
  title={pyannote. audio 2.1 speaker diarization pipeline: principle, benchmark, and recipe},
  author={Bredin, Herv{\'e}},
  booktitle={24th Interspeech Conference (INTERSPEECH 2023)},
  pages={1983--1987},
  year={2023},
  organization={ISCA}
}

@misc{bge-m3,
      title={BGE M3-Embedding: Multi-Lingual, Multi-Functionality, Multi-Granularity Text Embeddings Through Self-Knowledge Distillation}, 
      author={Jianlv Chen and Shitao Xiao and Peitian Zhang and Kun Luo and Defu Lian and Zheng Liu},
      year={2024},
      eprint={2402.03216},
      archivePrefix={arXiv},
      primaryClass={cs.CL}
}

@article{tonti2016rate,
  title={Rate of speech and emotional-cognitive regulation in the psychotherapeutic process: a pilot study},
  author={Tonti, Marco and Gelo, Omar CG},
  journal={Research in Psychotherapy: Psychopathology, Process and Outcome},
  volume={19},
  number={2},
  year={2016}
}

@article{fulmer2018using,
  title={Using psychological artificial intelligence (Tess) to relieve symptoms of depression and anxiety: randomized controlled trial},
  author={Fulmer, Russell and Joerin, Angela and Gentile, Breanna and Lakerink, Lysanne and Rauws, Michiel},
  journal={JMIR mental health},
  volume={5},
  number={4},
  pages={e9782},
  year={2018},
  publisher={JMIR Publications Inc., Toronto, Canada}
}

@article{rocco2018beyond,
  title={Beyond verbal behavior: An empirical analysis of speech rates in psychotherapy sessions},
  author={Rocco, Diego and Pastore, Massimiliano and Gennaro, Alessandro and Salvatore, Sergio and Cozzolino, Mauro and Scorza, Maristella},
  journal={Frontiers in psychology},
  volume={9},
  pages={978},
  year={2018},
  publisher={Frontiers Media SA}
}

@incollection{voss2024mental,
  title={Mental status examination},
  author={Voss, Rachel M and Das, Joe M},
  booktitle={StatPearls [Internet]},
  year={2024},
  publisher={StatPearls Publishing}
}

\appendix
\DefineVerbatimEnvironment{PromptBox}{Verbatim}{
  breaklines=true,
  breakanywhere=true,
  fontsize=\small,
  frame=single,
  framesep=3mm
}

\section{Source Recordings}
\label{app:source-recordings}

Table~\ref{tab:source-recordings} lists the 96 publicly available CBT educational recordings used to construct CBT-Audio.

\begingroup
\footnotesize
\setlength{\tabcolsep}{3pt}
\begin{longtable}{@{}>{\raggedright\arraybackslash}p{0.055\linewidth}>{\raggedright\arraybackslash}p{0.510\linewidth}>{\raggedright\arraybackslash}p{0.385\linewidth}@{}}
\caption{Source recordings used to construct CBT-Audio.}\label{tab:source-recordings}\\
\toprule
\textbf{ID} & \textbf{Recording title} & \textbf{Source URL} \\
\midrule
\endfirsthead
\caption[]{Source recordings used to construct CBT-Audio (continued).}\\
\toprule
\textbf{ID} & \textbf{Recording title} & \textbf{Source URL} \\
\midrule
\endhead
\midrule
\multicolumn{3}{r}{\emph{Continued on next page}}\\
\endfoot
\bottomrule
\endlastfoot
001 & Case study clinical example CBT: First session with a client with symptoms of depression (CBT model) & \url{https://www.youtube.com/watch?v=7LD8iC4NqXM} \\
002 & Case study clinical example: First session with a client with symptoms of social anxiety (CBT model) & \url{https://www.youtube.com/watch?v=XH2tF8oB3cw} \\
003 & Case study clinical example: Session with a client with Bipolar Disorder (fluctuations in mood) & \url{https://www.youtube.com/watch?v=mpE-oaix5kA} \\
004 & CBT for Social Anxiety Disorder: Using downward arrow and thought challenging techniques & \url{https://www.youtube.com/watch?v=W3hMmZQAdhw} \\
005 & Social Anxiety Disorder: CBT behavioural experiment case example & \url{https://www.youtube.com/watch?v=ExNs8o8A4fI} \\
006 & CBT formulation in anorexia: Case study clinical example & \url{https://www.youtube.com/watch?v=KqbXZJ80yFU} \\
007 & Case study clinical example: First session with a client with symptoms of depression (CBT model) & \url{https://www.youtube.com/watch?v=JKUFWK6iSsw} \\
008 & CBT Role-Play - Managing Anger & \url{https://www.youtube.com/watch?v=W_tDRu67JnI} \\
009 & CBT Role-Play - Depressive Symptoms and Lack of Motivation & \url{https://www.youtube.com/watch?v=8aDFvvjC6XM} \\
010 & Behavioral Therapy Counseling Role-Play - Client with Symptoms of Narcissistic Personality Disorder & \url{https://youtu.be/MyXdvt4_AZA?si=FWQjRUMrWiRVx8rS} \\
011 & CBT Counseling Role-Play - Clients with Symptoms of Borderline Personality Disorder & \url{https://www.youtube.com/watch?v=jQgkVKGqBCE} \\
012 & CBT Role Play - Catastrophizing and Decatastrophizing & \url{https://www.youtube.com/watch?v=nanU4vR993I} \\
013 & CBT Role Play - The Premack Principle & \url{https://www.youtube.com/watch?v=VKdjQYtbiic} \\
014 & CBT Role-Play - Anxiety and Counting Ritual & \url{https://www.youtube.com/watch?v=jlXmVqhaMds} \\
015 & CBT Role-Play - Anxiety and Guilt Related to Balancing Home and Work & \url{https://www.youtube.com/watch?v=osROod3Hmpg} \\
016 & CBT Role-Play - Anxiety Related to School Performance & \url{https://www.youtube.com/watch?v=KRDH89uP8wI} \\
017 & CBT Role-Play - Fear of Voyeuristic Video Recording & \url{https://www.youtube.com/watch?v=EO9F6b1R-ws} \\
018 & CBT Role-Play - Fear of Driving & \url{https://www.youtube.com/watch?v=ZSDJymlMOxg} \\
019 & CBT Role-Play - Test Anxiety & \url{https://www.youtube.com/watch?v=_pkXuC7vsl0} \\
020 & CBT Role-Play - Treating Heroin Use & \url{https://www.youtube.com/watch?v=PtCj7ecyHjg} \\
021 & CBT Role-Play - Cognitive Reframing an Experience of Emotional Abuse & \url{https://www.youtube.com/watch?v=CdyJ0iB_k00} \\
022 & CBT Role-Play - Dating Anxiety & \url{https://www.youtube.com/watch?v=Xj3q96mCfC8} \\
023 & CBT Role-Play - Anxiety Related to Being Away from Home & \url{https://www.youtube.com/watch?v=k_78UKFERGE} \\
024 & CBT Role-Play - Challenging Relationship with Coworker & \url{https://www.youtube.com/watch?v=Pb1b2D26QsQ} \\
025 & CBT Role-Play - Challenging Relationship with Family Member & \url{https://www.youtube.com/watch?v=XbYGdo9lMsQ} \\
026 & CBT Role-Play - Loss of Hope & \url{https://www.youtube.com/watch?v=pllei-yDO8c} \\
027 & CBT Role-Play - The Premack Principle with Weight Loss & \url{https://www.youtube.com/watch?v=vdMnsJJvqR4} \\
028 & CBT Role-Play -  Problem Solving and Decatastrophizing after Job Loss & \url{https://www.youtube.com/watch?v=YpgvoXv3o-w} \\
029 & CBT Role-Play - Behavioral Activation and Depression & \url{https://www.youtube.com/watch?v=sDrakgSYvzc} \\
030 & CBT Role-Play - Subjective Units of Distress, Anxiety, and Public Speaking & \url{https://www.youtube.com/watch?v=6Hcebt_wBVw} \\
031 & CBT Role-Play - Exposure and Response Prevention - Early Session & \url{https://www.youtube.com/watch?v=TMVRjab1TmE} \\
032 & CBT Role-Play - Exposure and Response Prevention - Later Session & \url{https://www.youtube.com/watch?v=yFcqj_ml8dc} \\
033 & CBT Role-Play - Behavioral Activation and Postpartum Depression & \url{https://www.youtube.com/watch?v=SNstOn6owcI} \\
034 & CBT Role-Play - Anxiety Related to Dishonesty in Social Situations & \url{https://www.youtube.com/watch?v=LmB3ZQ2F1MY} \\
035 & CBT Role-Play – Subjective Units of Distress - Fear of Driving & \url{https://www.youtube.com/watch?v=iQvmMBi-XvE} \\
036 & CBT Role-Play - Downward Arrow Technique & \url{https://www.youtube.com/watch?v=Wx8F9uwQTnY} \\
037 & CBT Role-Play - Subjective Units of Distress - Dog Phobia & \url{https://www.youtube.com/watch?v=O8dpj-Hqf0g} \\
038 & CBT Role-Play - Mixed Feelings about Dating Someone Older & \url{https://www.youtube.com/watch?v=k_Rw-bpRe-I} \\
039 & CBT Role-Play – Complete Session – Social Anxiety Disorder – Part 1 & \url{https://www.youtube.com/watch?v=gbBn8EzZx3w} \\
040 & CBT Role-Play – Complete Session – Social Anxiety Disorder – Part 2 & \url{https://www.youtube.com/watch?v=8K4HW6_MvoU} \\
041 & CBT Role-Play – Complete Session – Social Anxiety Disorder – Part 3 & \url{https://www.youtube.com/watch?v=jmwQ3SE6Uew} \\
042 & CBT Role-Play – Complete Session – Social Anxiety Disorder – Part 4 & \url{https://www.youtube.com/watch?v=LuuKIF4-F_Q} \\
043 & CBT Role-Play – Complete Session – Social Anxiety Disorder – Part 5 & \url{https://www.youtube.com/watch?v=o2Cv6Mlp3KY} \\
044 & CBT Role-Play – Complete Session – Social Anxiety Disorder – Part 6 & \url{https://www.youtube.com/watch?v=TcewFGydzPM} \\
045 & CBT Role-Play – Complete Session – Depression and Alcohol Use – Part 1 & \url{https://www.youtube.com/watch?v=9cYeZdKeXsc} \\
046 & CBT Role-Play – Complete Session – Depression and Alcohol Use – Part 2 & \url{https://www.youtube.com/watch?v=j1mehpweq4g} \\
047 & CBT Role-Play – Complete Session – Depression and Alcohol Use – Part 3 & \url{https://www.youtube.com/watch?v=R9dIc-inEGw} \\
048 & CBT Role-Play – Complete Session – Depression and Alcohol Use – Part 4 & \url{https://www.youtube.com/watch?v=RHjCPoWnDLs} \\
049 & CBT Role-Play – Complete Session – Depression and Alcohol Use – Part 5 & \url{https://www.youtube.com/watch?v=uYD36n3xM7E} \\
050 & CBT Role-Play – Complete Session – Depression and Alcohol Use – Part 6 & \url{https://www.youtube.com/watch?v=ZDpMz864zBA} \\
051 & CBT Role-Play – Complete Session – Depression and Alcohol Use – Part 7 & \url{https://www.youtube.com/watch?v=bdHut6b8-2Y} \\
052 & CBT Role-Play – Complete Session – Depression and Alcohol Use – Part 8 & \url{https://www.youtube.com/watch?v=cGCE4rxRhZA} \\
053 & CBT Role-Play – Complete Session – Depression and Alcohol Use – Part 9 & \url{https://www.youtube.com/watch?v=ccMJbdHC6cg} \\
054 & CBT Role-Play – Complete Session – Depression and Alcohol Use – Part 10 & \url{https://www.youtube.com/watch?v=dde6B4lpaLM} \\
055 & CBT Role-Play – Complete Session – Low Self-Confidence at Work – Part 1 & \url{https://www.youtube.com/watch?v=tq_yyMVax_c} \\
056 & CBT Role-Play – Complete Session – Low Self-Confidence at Work – Part 2 & \url{https://www.youtube.com/watch?v=G40ywooLnns} \\
057 & CBT Role-Play – Complete Session – Low Self-Confidence at Work – Part 3 & \url{https://www.youtube.com/watch?v=DcAqhHb5-hk} \\
058 & CBT Role-Play – Complete Session – Low Self-Confidence at Work – Part 4 & \url{https://www.youtube.com/watch?v=U6VzVpsqG40} \\
059 & CBT Role-Play – Complete Session – Low Self-Confidence at Work – Part 5 & \url{https://www.youtube.com/watch?v=-tbKuW6k1cE} \\
060 & CBT Role-Play – Complete Session – Low Self-Confidence at Work – Part 6 & \url{https://www.youtube.com/watch?v=KuHLL2AE-SE} \\
061 & CBT Role-Play – Complete Session – Low Self-Confidence at Work – Part 7 & \url{https://www.youtube.com/watch?v=jS1KE3_Pqlc} \\
062 & CBT Role-Play – Complete Session – Low Self-Confidence at Work – Part 8 & \url{https://www.youtube.com/watch?v=Ac0aZgha6Fk} \\
063 & CBT Role-Play – Complete Session – Low Self-Confidence at Work – Part 9 & \url{https://www.youtube.com/watch?v=Q_kYJ63RUAA} \\
064 & CBT Role-Play – Complete Session – Low Self-Confidence at Work – Part 10 & \url{https://www.youtube.com/watch?v=C4kdF-3btbg} \\
065 & CBT Role-Play – Complete Session – Low Self-Confidence at Work – Part 11 & \url{https://www.youtube.com/watch?v=Bub8YLl1fUI} \\
066 & CBT Role-Play – Complete Session – Low Self-Confidence at Work – Part 12 & \url{https://www.youtube.com/watch?v=rRze7Na1MXg} \\
067 & CBT Role-Play – Complete Session – Anxiety, Obsessions, and Compulsions – Part 1 & \url{https://www.youtube.com/watch?v=rs9wsvkf7uQ} \\
068 & CBT Role-Play – Complete Session – Anxiety, Obsessions, and Compulsions – Part 2 & \url{https://www.youtube.com/watch?v=JfgNaHtoK8M} \\
069 & CBT Role-Play – Complete Session – Anxiety, Obsessions, and Compulsions – Part 3 & \url{https://www.youtube.com/watch?v=qAXQ9hlf42Y} \\
070 & CBT Role-Play – Complete Session – Anxiety, Obsessions, and Compulsions – Part 4 & \url{https://www.youtube.com/watch?v=GTe_a7HJxqg} \\
071 & CBT Role-Play – Complete Session – Anxiety, Obsessions, and Compulsions – Part 5 & \url{https://www.youtube.com/watch?v=TpStHveF3wo} \\
072 & CBT Role-Play – Complete Session – Anxiety, Obsessions, and Compulsions – Part 6 & \url{https://www.youtube.com/watch?v=W-fNT9mJuAA} \\
073 & CBT Role-Play – Complete Session – Anxiety, Obsessions, and Compulsions – Part 7 & \url{https://www.youtube.com/watch?v=jZ-nv6SUCEs} \\
074 & CBT Role Play – Complete Session – Anxiety, Obsessions, and Compulsions – Part 8 & \url{https://www.youtube.com/watch?v=Y8GdmE1dfSA} \\
075 & CBT Role-Play – Complete Session – Anxiety, Obsessions, and Compulsions – Part 9 & \url{https://www.youtube.com/watch?v=Z2TTSGvNhUs} \\
076 & CBT Role-Play – Complete Session – Anxiety, Obsessions, and Compulsions – Part 10 & \url{https://www.youtube.com/watch?v=uj-jMur6xHI} \\
077 & CBT Role-Play - Fear of Driving Over Bridges & \url{https://www.youtube.com/watch?v=tBd5VGCKnjA} \\
078 & Sample CBT Session on Social Anxiety & \url{https://www.youtube.com/watch?v=jfVeeGJHFjA} \\
079 & CBT for PTSD: Example of how grounding techniques can be used in therapy & \url{https://www.youtube.com/watch?v=RybY4zIecQ4} \\
080 & CBT Demo - Addressing Suicidality & \url{https://www.youtube.com/watch?v=AG15pPEStK0} \\
081 & CBT Demo Socratic Questioning & \url{https://www.youtube.com/watch?v=sW5HDbm09ZE} \\
082 & CBT Demo Grounding Techniques 54321 & \url{https://www.youtube.com/watch?v=gL93G4IBWZQ} \\
083 & Role Play: Cognitive Behaviour Therapy & \url{https://www.youtube.com/watch?v=x7HJmVx3qN4} \\
084 & CBT and Gestalt Integration Therapy Role-Play - Grounding and Awareness Techniques & \url{https://www.youtube.com/watch?v=ex0uc6r86Qw} \\
085 & REBT Role-Play - Understanding the Difference Between Self-Judgment and the Evaluation of Behaviors & \url{https://www.youtube.com/watch?v=L4oDAYZWLhs} \\
086 & Rational Emotive Behavior Therapy (REBT) Role-Play - Complicated Grief & \url{https://www.youtube.com/watch?v=pGxc8pyhV2Q} \\
087 & REBT Role-Play - Addressing Belief Regarding Expectation of Respect & \url{https://www.youtube.com/watch?v=DPOFpRQqInw} \\
088 & REBT Role-Play - Postponing Gratification & \url{https://www.youtube.com/watch?v=RsI7Iycy5H4} \\
089 & CBT Demo Exposure Response Prevention & \url{https://www.youtube.com/watch?v=hAvkumT_Vzw} \\
090 & CBT Demo Exposure Response Prevention Part 2 & \url{https://www.youtube.com/watch?v=1ISWXhNnx1o} \\
091 & CBT Demo The Sleep Diary & \url{https://www.youtube.com/watch?v=4RA7d7P3bBs} \\
092 & CBT Demo Dealing with Uncertainty & \url{https://www.youtube.com/watch?v=7vrM08pydug} \\
093 & CBT Demo - Recognizing Anger Cues & \url{https://www.youtube.com/watch?v=grTGpx30gc0} \\
094 & CBT Demo Addictive Beliefs & \url{https://www.youtube.com/watch?v=YfY58kxQi1U} \\
095 & CBT Demo Thought Record Part 1 Identifying Hot Thoughts & \url{https://www.youtube.com/watch?v=fqgsMJsSMeQ} \\
096 & Full-Length CBT Session Demo & \url{https://www.youtube.com/watch?v=mx_vC45Fmw4} \\
\end{longtable}
\endgroup

\section{Reference Label Generation}
\label{app:reference-label-generation}

This section provides the full prompts and additional analyses for reference label construction.

\subsection{Prompt for Emotional-State Description Generation}
\label{app:ssr-description-prompt}
\paragraph{User prompt template for description generation.}
\begin{PromptBox}
You are analyzing a therapy session. You will receive:
1. TEXT context (previous conversation turns) - to understand the topic being discussed
2. AUDIO clip (patient's current turn) - to assess their emotional state

Your task: Based on the AUDIO, describe how the patient FEELS emotionally.
Use the audio (tone, pace, voice quality) as evidence, but describe their emotional state, not just how they sound.

The context is provided only to help you understand the conversation topic.

Examples of good descriptions:
- "The patient feels anxious and stressed, struggling to maintain composure."
- "The patient feels calm and at ease, with no signs of emotional distress."

---

Previous conversation:
{context}

Now listen to the patient's response and describe how they feel emotionally.

Output ONLY a 1-2 sentence description of their emotional state. No preamble or explanation.
\end{PromptBox}

\paragraph{Description-to-label mapping.} After the model generated a 1--2 sentence emotional-state description, the description was embedded using \texttt{BAAI/bge-m3}. The embedding was compared with the five anchor-description embeddings in Table~\ref{tab:anchors} using cosine similarity. The final SSR score for that run was assigned as the label of the closest anchor.

\paragraph{Model and decoding settings.}
Table~\ref{tab:ssr-label-settings} summarises the settings used to generate reference-label descriptions before SSR matching. For each retained patient-side turn, the input consisted of the previous three conversation turns as text context and the current patient turn as audio. No separate system prompt was used; the instruction below was passed as the user text prompt together with the audio clip.

\begin{table}[ht]
\centering
\small
\caption{Settings used for SSR-based reference label generation.}
\label{tab:ssr-label-settings}
\begin{tabularx}{\linewidth}{@{}lX@{}}
\toprule
\textbf{Setting} & \textbf{Value} \\
\midrule
Description-generation model & \texttt{gpt-audio-1.5} \\
Embedding model for SSR matching & \texttt{BAAI/bge-m3} \\
Number of independent runs & 3 \\
Temperature & 0.7 \\
Maximum output length & 4096 tokens \\
Patient-turn filter & Patient-side turns with duration $\geq$2.0 seconds \\
Text context window & Previous 3 conversation turns \\
Aggregation rule & Rounded median across the three SSR scores; this is equivalent to majority voting when at least two runs agree and gives the middle ordinal score when all three runs differ. \\
\bottomrule
\end{tabularx}
\end{table}

\subsection{Label Stability Across Three Independent Runs}
\label{app:run-distribution}

\begin{figure}[ht]
\centering
\includegraphics[width=0.9\linewidth]{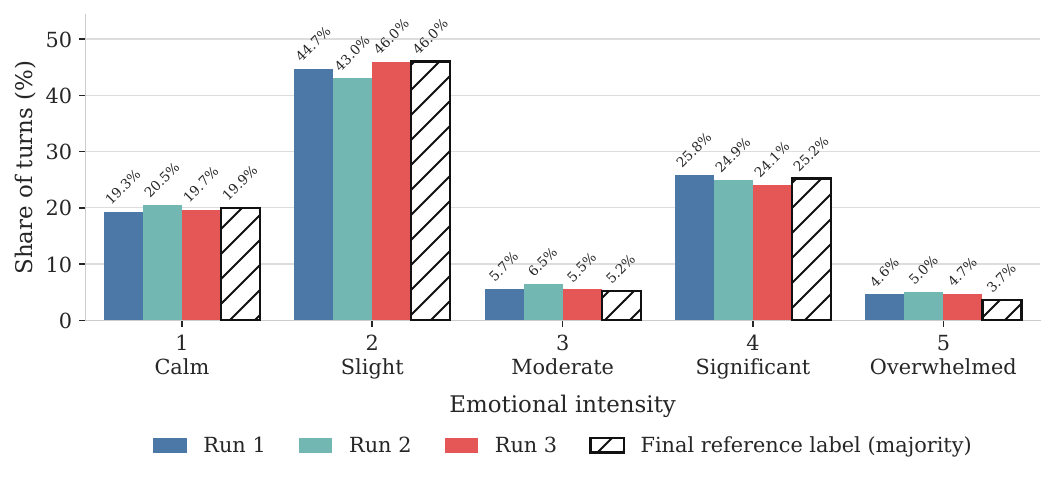}
\caption{Label distribution across three independent GPT-audio-1.5 SSR runs and the final aggregated SSR label. The final label was computed as the rounded median across the three run-level SSR scores.}
\label{fig:three-run-majority-distribution}
\end{figure}

To assess whether the SSR labeling procedure was stable across repeated GPT-audio calls, we ran the full reference-label generation pipeline three times independently. Figure~\ref{fig:three-run-majority-distribution} compares the score distributions from the three runs with the final aggregated SSR label. The three runs produce similar distributions.

\subsection{SSR Labels Compared with Direct Numeric Prompting}
\label{app:ssr-vs-numeric}

\begin{figure}[ht]
\centering
\includegraphics[width=0.85\linewidth]{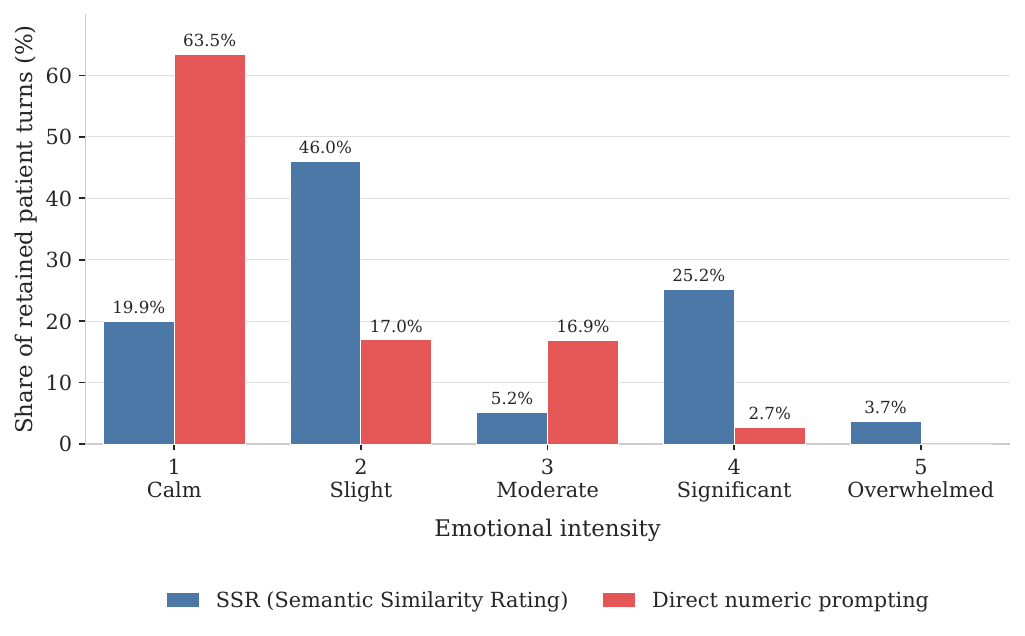}%
\caption{Comparison between SSR-based labels and direct numeric prompting.}
\label{fig:ssr-vs-numeric-distribution}
\end{figure}

We compared SSR with direct numeric prompting, where GPT-audio was asked to output a score directly on the 1--5 scale, given the anchor description. As shown in Figure~\ref{fig:ssr-vs-numeric-distribution}, direct numeric prompting produced a highly concentrated distribution, especially at the lowest distress level. In contrast, SSR produced a broader distribution across the five levels, which better supports the evaluation of model predictions.

\section{Human Annotation Validation}
\label{app:human-validation}

We conducted an expert validation study on 194 retained patient turns sampled from eight source recordings: 001, 002, 004, 005, 006, 007, 009, and 016. Five expert annotators independently rated patient distress intensity using the same five-level ordinal scale used for the SSR reference labels. One annotator completed 178 turns, while the other four annotators completed all 194 turns.

\subsection{Inter-rater agreement}
\label{app:inter-rater-agreement}

Table~\ref{tab:human-human-agreement} summarizes agreement among the five expert annotators. We report ordinal Krippendorff's $\alpha$ as a group-level agreement metric and mean pairwise quadratic-weighted Cohen's $\kappa$ (QWK) as a pairwise ordinal agreement metric. The results indicate moderate inter-rater agreement, which is expected when rating patient distress from speech clips.

\begin{table}[!htbp]
\centering
\small
\caption{Human inter-annotator agreement on the expert-labeled validation subset. Pairwise metrics are averaged over the ten annotator pairs. QWK = quadratic-weighted Cohen's $\kappa$.}
\label{tab:human-human-agreement}
\begin{tabular}{lccccc}
\toprule
Subset & Annotators & Clips & Krippendorff's $\alpha$ & Mean QWK & Mean exact match \\
\midrule
Expert-labeled subset & 5 & 194 & 0.439 & 0.441 & 50.4\% \\
\bottomrule
\end{tabular}
\end{table}

\subsection{Human consensus compared with SSR reference labels}
\label{app:human-vs-reference}

We next compared the SSR reference labels against individual expert annotations and against the per-clip human consensus label. The consensus label was computed as the median of the available expert ratings for each clip. Because the labels are ordinal, a one-level difference still indicates close agreement. We therefore report both MAE and within-one-level agreement.

As shown in Table~\ref{tab:human-reference-agreement}, the human consensus label was within one ordinal level of the SSR reference label for 81.4\% of clips. The positive Spearman correlation further indicates that clips assigned higher distress by human consensus also tend to receive higher SSR reference labels.

\begin{table}[!htbp]
\centering
\small
\caption{Comparison between expert annotations and SSR reference labels on the validation subset. The consensus label is the per-clip median of the available expert annotations.}
\label{tab:human-reference-agreement}
\begin{tabular}{lcccc}
\toprule
Comparison & $N$ & MAE & Within $\pm$1 & Spearman $\rho$ \\
\midrule
Human consensus & 194 & 0.856 & 81.4\% & 0.434 \\
\midrule
Expert 1 & 178 & 0.854 & 82.6\% & 0.527 \\
Expert 2 & 194 & 0.918 & 77.3\% & 0.298 \\
Expert 3 & 194 & 0.881 & 80.9\% & 0.388 \\
Expert 4 & 194 & 0.902 & 76.8\% & 0.392 \\
Expert 5 & 194 & 0.918 & 80.4\% & 0.248 \\
\bottomrule
\end{tabular}
\end{table}

\paragraph{Model-ranking robustness.}
Beyond clip-level agreement, we also examined whether model rankings are stable when evaluated using the full SSR reference labels versus the expert-labeled subset. The resulting rank correlation was high (Spearman's $\rho = 0.882$, $p < 0.001$), suggesting that the SSR reference labels preserve the relative ordering of model performance observed on the expert-labeled subset.

\section{Evaluation Prompts and Inference Settings}
\label{app:evaluation-prompts}

All evaluated models used the same five-level distress-intensity scale and the same conversation-context setting. Each input included up to three previous conversation turns as context. Generation used \texttt{max\_new\_tokens=64}, \texttt{temperature=0.7}, \texttt{top\_p=0.9}, and sampling enabled. For transcript-only evaluation, some models required an audio input field in their inference stage. To keep the input format valid while removing acoustic information, we supplied a one-second silent audio file.

\paragraph{System prompt.}
\begin{Verbatim}[breaklines=true,breakanywhere=true,fontsize=\small]
You are an expert in clinical psychology specialising in Cognitive Behavioral Therapy (CBT).
\end{Verbatim}

\subsection{Audio-only prompt}
\label{app:prompt-ao}

\begin{Verbatim}[breaklines=true,breakanywhere=true,fontsize=\small]
Listen to this audio clip of a patient speaking during a CBT therapy session.

Context from the conversation:
{context}

Based on the patient's audio, what they said and how they said it, rate their emotional intensity on a scale of 1 to 5:
1 = Calm, at ease, with no emotional distress
2 = Slightly uneasy, with mild emotional activation
3 = Moderately distressed, with clear emotional activation
4 = Significantly distressed, struggling to maintain composure
5 = Overwhelmed with intense emotional distress

Respond with ONLY a single number (1-5).
\end{Verbatim}

\subsection{Audio+transcript prompt}
\label{app:prompt-at}

\begin{Verbatim}[breaklines=true,breakanywhere=true,fontsize=\small]
Listen to this audio clip and read the transcript of a patient speaking during a CBT therapy session.

Context from the conversation:
{context}

Patient's current statement (transcript):
"{patient_text}"

Based on the patient's audio and transcript, both what they said and how they said it, rate their emotional intensity on a scale of 1 to 5:
1 = Calm, at ease, with no emotional distress
2 = Slightly uneasy, with mild emotional activation
3 = Moderately distressed, with clear emotional activation
4 = Significantly distressed, struggling to maintain composure
5 = Overwhelmed with intense emotional distress

Respond with ONLY a single number (1-5).
\end{Verbatim}

\subsection{Transcript-only prompt}
\label{app:prompt-to}

\begin{Verbatim}[breaklines=true,breakanywhere=true,fontsize=\small]
Read the following transcript of a patient speaking during a CBT therapy session.

Context from the conversation:
{context}

Patient's current statement:
"{patient_text}"

Based on what the patient said, rate their emotional intensity on a scale of 1 to 5:
1 = Calm, at ease, with no emotional distress
2 = Slightly uneasy, with mild emotional activation
3 = Moderately distressed, with clear emotional activation
4 = Significantly distressed, struggling to maintain composure
5 = Overwhelmed with intense emotional distress

Respond with ONLY a single number (1-5).
\end{Verbatim}

\paragraph{Compute resources.}
All model inference was run on a local workstation equipped with an NVIDIA RTX A6000 GPU with 48GB VRAM. No model training or fine-tuning was performed, the experiments used inference only.

\section{Model Details}
\label{app:model-details}

Table~\ref{tab:model-overview} summarizes the open-source audio language models evaluated in CBT-Audio. We report the main audio pathway and the interface used to connect audio representations to the language model.

\begin{table}[ht]
\centering
\small
\caption{Overview of the open-source audio language models evaluated in CBT-Audio.}
\label{tab:model-overview}
\begin{tabularx}{\linewidth}{@{}p{0.20\linewidth}p{0.34\linewidth}X@{}}
\toprule
\textbf{Model} & \textbf{Core audio pathway} & \textbf{LLM backbone / audio interface} \\
\midrule
SALMONN-7B 
& Whisper + BEATs 
& Window-level Q-Former maps audio features into the LLM input space; Vicuna backbone with LoRA adaptation \\

MiniCPM-o 2.6 
& Whisper-medium-300M audio encoder 
& Audio representations are projected into a Qwen2.5-7B-Instruct model; speech decoder is not used for our text-output evaluation \\

Phi-4-Multimodal 
& Custom Conformer-based audio encoder + audio projector 
& Frozen Phi-4-Mini backbone with modality-specific speech/audio LoRA adapter \\

Qwen2.5-Omni-7B 
& Whisper-large-v3-based audio encoder 
& Audio, text, and other modality tokens are processed by the Thinker LLM; Talker speech generation module is not used for our text-output evaluation \\

Qwen3-Omni-30B-A3B 
& Audio Transformer (AuT) encoder 
& Thinker--Talker MoE architecture based on Qwen3-30B-A3B; text output is produced by the Thinker \\

Kimi-Audio-7B 
& Discrete semantic tokens + continuous Whisper features 
& Qwen2.5-7B-initialized Audio LLM with shared layers and parallel text/audio output heads \\

Ultravox-v0.5-Llama-3.1-8B
& Whisper-large-v3-turbo encoder 
& Multimodal projector maps audio features into a frozen Llama backbone \\

Audio Flamingo 3 
& AF-Whisper audio encoder 
& MLP audio adaptor maps AF-Whisper features into a Qwen2.5-7B LLM backbone \\

Voxtral-Mini 
& Whisper-large-v3 encoder 
& Adapter/downsampling layer maps audio features into a Ministral-3B language decoder \\

Gemma-4-E4B 
& USM-style Conformer audio encoder 
& Native Gemma-4 E4B multimodal decoder with audio input support \\
\bottomrule
\end{tabularx}
\end{table}

\begin{landscape}
\section{Full Model Results}
\label{app:full-results}
This section report the full model-level results for all models and all input conditions, including confidence intervals for each metric.
\begingroup
\scriptsize
\setlength{\tabcolsep}{2.2pt}
\renewcommand{\arraystretch}{1.08}
\begin{longtable}{@{}>{\raggedright\arraybackslash}p{0.20\linewidth}
>{\centering\arraybackslash}p{0.14\linewidth}
>{\centering\arraybackslash}p{0.14\linewidth}
>{\centering\arraybackslash}p{0.14\linewidth}
>{\centering\arraybackslash}p{0.14\linewidth}
>{\centering\arraybackslash}p{0.14\linewidth}@{}}
\caption{Full model results on CBT-Audio. Lower is better for MAE and RMSE; higher is better for Spearman's $\rho$, QWK, and $A_1$.}\\
\label{tab:full-model-results}\\
\toprule
\textbf{Model condition} & \textbf{MAE} & \textbf{$A_1$ (\%)} & \textbf{Spearman $\rho$} & \textbf{QWK} & \textbf{RMSE} \\
\midrule
\endfirsthead

\caption[]{Full model results on CBT-Audio (continued).}\\
\toprule
\textbf{Model condition} & \textbf{MAE} & \textbf{$A_1$ (\%)} & \textbf{Spearman $\rho$} & \textbf{QWK} & \textbf{RMSE} \\
\midrule
\endhead

\midrule
\multicolumn{6}{r}{\emph{Continued on next page}}\\
\endfoot

\bottomrule
\endlastfoot
Gemma-4-E4B (AT) & 0.720 [0.691, 0.751] & 90.0 [88.6, 91.3] & 0.587 [0.554, 0.616] & 0.535 [0.502, 0.564] & 0.970 [0.940, 0.999] \\
Gemma-4-E4B (TO) & 0.746 [0.716, 0.780] & 89.0 [87.5, 90.4] & 0.578 [0.545, 0.610] & 0.530 [0.498, 0.561] & 1.001 [0.970, 1.034] \\
Qwen3-Omni-30B-A3B (AO) & 0.785 [0.757, 0.812] & 89.9 [88.6, 91.3] & 0.583 [0.551, 0.615] & 0.519 [0.489, 0.548] & 1.001 [0.973, 1.028] \\
Qwen3-Omni-30B-A3B (AT) & 0.813 [0.785, 0.840] & 88.2 [86.7, 89.5] & 0.583 [0.550, 0.614] & 0.500 [0.470, 0.529] & 1.030 [1.003, 1.056] \\
Gemma-4-E4B (AO) & 0.824 [0.792, 0.856] & 86.5 [84.9, 88.1] & 0.483 [0.446, 0.519] & 0.434 [0.400, 0.466] & 1.066 [1.034, 1.100] \\
Qwen3-Omni-30B-A3B (TO) & 0.873 [0.843, 0.903] & 84.2 [82.5, 85.8] & 0.531 [0.496, 0.563] & 0.455 [0.423, 0.486] & 1.108 [1.078, 1.139] \\
Ultravox-v0.5 (TO) & 0.883 [0.853, 0.917] & 82.0 [80.3, 83.6] & 0.524 [0.488, 0.558] & 0.428 [0.397, 0.456] & 1.133 [1.103, 1.164] \\
Ultravox-v0.5 (AT) & 0.895 [0.862, 0.926] & 81.4 [79.7, 83.2] & 0.513 [0.478, 0.553] & 0.413 [0.382, 0.445] & 1.148 [1.114, 1.179] \\
Ultravox-v0.5 (AO) & 0.912 [0.878, 0.943] & 80.0 [78.3, 81.8] & 0.419 [0.382, 0.459] & 0.358 [0.327, 0.394] & 1.167 [1.134, 1.199] \\
Qwen2.5-Omni-7B (AT) & 0.931 [0.897, 0.962] & 80.6 [78.9, 82.3] & 0.464 [0.427, 0.500] & 0.383 [0.350, 0.415] & 1.158 [1.128, 1.186] \\
Qwen2.5-Omni-7B (TO) & 0.941 [0.908, 0.973] & 79.8 [78.0, 81.6] & 0.463 [0.425, 0.497] & 0.387 [0.354, 0.419] & 1.175 [1.144, 1.205] \\
Qwen2.5-Omni-7B (AO) & 0.963 [0.932, 0.996] & 80.5 [78.7, 82.3] & 0.431 [0.390, 0.465] & 0.356 [0.321, 0.387] & 1.193 [1.161, 1.225] \\
Kimi-Audio-7B (AT) & 0.987 [0.957, 1.016] & 81.4 [79.7, 83.1] & 0.339 [0.303, 0.381] & 0.254 [0.225, 0.288] & 1.179 [1.150, 1.205] \\
Kimi-Audio-7B (AO) & 1.062 [1.030, 1.091] & 78.1 [76.2, 80.0] & 0.261 [0.221, 0.303] & 0.187 [0.157, 0.218] & 1.263 [1.229, 1.293] \\
MiniCPM-o 2.6 (AT) & 1.079 [1.046, 1.109] & 72.9 [70.9, 74.9] & 0.503 [0.471, 0.541] & 0.280 [0.256, 0.303] & 1.279 [1.251, 1.306] \\
MiniCPM-o 2.6 (TO) & 1.089 [1.057, 1.121] & 70.4 [68.3, 72.4] & 0.528 [0.493, 0.561] & 0.293 [0.269, 0.317] & 1.312 [1.283, 1.340] \\
Voxtral-Mini (AT) & 1.104 [1.073, 1.132] & 74.9 [72.8, 76.9] & 0.372 [0.335, 0.411] & 0.210 [0.187, 0.236] & 1.276 [1.248, 1.303] \\
Voxtral-Mini (AO) & 1.106 [1.077, 1.135] & 75.7 [73.8, 77.6] & 0.279 [0.237, 0.321] & 0.168 [0.141, 0.195] & 1.282 [1.254, 1.309] \\
Audio Flamingo 3 (AT) & 1.106 [1.073, 1.137] & 71.8 [69.9, 73.9] & 0.410 [0.371, 0.448] & 0.238 [0.211, 0.263] & 1.308 [1.280, 1.336] \\
Phi-4-Multimodal (AO) & 1.114 [1.080, 1.150] & 71.1 [69.0, 73.1] & 0.278 [0.234, 0.322] & 0.211 [0.178, 0.245] & 1.352 [1.320, 1.386] \\
Voxtral-Mini (TO) & 1.115 [1.087, 1.145] & 73.1 [71.1, 75.1] & 0.353 [0.313, 0.392] & 0.203 [0.176, 0.228] & 1.304 [1.275, 1.332] \\
MiniCPM-o 2.6 (AO) & 1.129 [1.097, 1.160] & 70.2 [68.1, 72.4] & 0.394 [0.352, 0.437] & 0.212 [0.189, 0.236] & 1.336 [1.305, 1.364] \\
Kimi-Audio-7B (TO) & 1.133 [1.098, 1.169] & 69.6 [67.6, 71.7] & 0.387 [0.350, 0.427] & 0.265 [0.237, 0.296] & 1.391 [1.354, 1.426] \\
Phi-4-Multimodal (AT) & 1.137 [1.098, 1.170] & 68.8 [66.9, 70.9] & 0.359 [0.319, 0.399] & 0.250 [0.220, 0.279] & 1.402 [1.366, 1.434] \\
Audio Flamingo 3 (TO) & 1.143 [1.113, 1.173] & 71.3 [69.4, 73.3] & 0.328 [0.287, 0.366] & 0.186 [0.162, 0.212] & 1.343 [1.313, 1.373] \\
Phi-4-Multimodal (TO) & 1.156 [1.117, 1.194] & 65.9 [63.7, 68.3] & 0.357 [0.317, 0.399] & 0.242 [0.214, 0.273] & 1.427 [1.390, 1.463] \\
Audio Flamingo 3 (AO) & 1.156 [1.124, 1.189] & 69.8 [67.7, 71.9] & 0.303 [0.262, 0.344] & 0.168 [0.144, 0.193] & 1.359 [1.329, 1.389] \\
SALMONN-7B (TO) & 1.585 [1.539, 1.629] & 35.4 [33.4, 37.4] & 0.082 [0.040, 0.124] & 0.010 [0.005, 0.015] & 1.907 [1.872, 1.940] \\
SALMONN-7B (AT) & 1.601 [1.553, 1.644] & 34.4 [32.4, 36.5] & 0.038 [-0.009, 0.084] & 0.002 [-0.003, 0.005] & 1.922 [1.887, 1.957] \\
SALMONN-7B (AO) & 1.606 [1.559, 1.650] & 34.2 [32.2, 36.2] & 0.019 [-0.039, 0.070] & -0.001 [-0.009, 0.003] & 1.929 [1.894, 1.964] \\
\end{longtable}
\endgroup
\end{landscape}

\clearpage
\begin{landscape}

\section{Full Case Studies}
\label{app:full-case-studies}

Table~\ref{tab:full-case-studies} provides the full context and target patient turns for the case studies summarized in Figure~\ref{fig:casestudy}. Each context contains the three preceding turns used during model evaluation.

\begingroup
\footnotesize
\setlength{\tabcolsep}{4pt}
\renewcommand{\arraystretch}{1.12}
\begin{longtable}{@{}>{\raggedright\arraybackslash}p{0.08\linewidth}>{\raggedright\arraybackslash}p{0.55\linewidth}>{\raggedright\arraybackslash}p{0.29\linewidth}>{\centering\arraybackslash}p{0.04\linewidth}@{}}
\caption{Full version of the case studies summarized in Figure~\ref{fig:casestudy}. Ref. is the SSR reference label on the 1--5 distress-intensity scale.}\label{tab:full-case-studies}\\
\toprule
\textbf{Case} & \textbf{Full context} & \textbf{Patient transcript} & \textbf{Ref.} \\
\midrule
\endfirsthead

\caption[]{Full version of the case studies summarized in Figure~\ref{fig:casestudy} (continued).}\\
\toprule
\textbf{Case} & \textbf{Full context} & \textbf{Patient transcript} & \textbf{Ref.} \\
\midrule
\endhead

\midrule
\multicolumn{4}{r}{\emph{Continued on next page}}\\
\endfoot

\bottomrule
\endlastfoot

\textbf{A}\par{\scriptsize Video 001 Turn 040}
&
\textbf{Therapist:} Okay. Okay, so it sounds like quite a lot has been happening. You've been feeling very low. You've had uni pressure on. You've had problems with your parents' marriage. And it sounds like you feel like you can't really reach out to anyone, that you can't understand, but you can't really reach out to your parents right now. You can't really confide in your friends. So I suppose I'm just wondering if you can tell me a bit more about how you've been feeling, I mean, say this past week.\par \textbf{Patient:} Really not good, to be honest. I actually... If I don't have to, then I don't really leave the house or get out of bed. I can't find the motivation for the things I used to enjoy. I used to love doing sports or going out and now I just prefer to lie in bed and not really do anything. I missed a few lectures this week that I should have gone to.\par \textbf{Therapist:} Right, so I suppose I'm just wondering then, let's see if we can think about one thing, say a lecture. What kind of thoughts do you have about going to that lecture before it happens?
&
\emph{I don't really see the point anymore. If I can't do as well as I should be doing, then what's the point in putting myself in those positions? I just, I can't be bothered anymore to try. It just stresses me out more than I need to.}
&
5 \\
\midrule
\textbf{B}\par{\scriptsize Video 043 Turn 005}
&
\textbf{Therapist:} Right. Let's actually go right before the moment. So you're still in the, you're still sitting down and you're getting ready to present. Let's just say you're second to present. Yeah. Like we've talked about. And you're thinking they're going to see me as anxious. They're going to think I don't belong here. So these thoughts occur because you have an expectation, a rule, an attitude that you brought into that class. When you walked into that class, you have these expectations already. It's perfectly natural. It's not right or wrong. Everyone has these beliefs. And whatever these beliefs are, and we'll talk about that at another point, they combine with the stressor of having this presentation in front of you and the classmates watching you. And they result in these thoughts. We call them automatic thoughts. And just like the beliefs, they're not good or bad. They don't say anything about the person having them. They're just thoughts. Actually, they're fairly understandable and natural thoughts. The problem is that they lead to symptoms. So you're sitting there at your desk, and you're thinking, they're going to see that I'm anxious, they're going to realize I'm not good enough or think I'm not good enough.\par \textbf{Patient:} Right.\par \textbf{Therapist:} What reaction do you have when you have that thought?
&
\emph{I guess I'm like very observant, like I start looking around like I'm sweating, like I'm having physical responses and I'm looking around to see if anyone's already looking at me, like hypervigilance.}
&
2 \\
\midrule
\textbf{C}\par{\scriptsize Video 023 Turn 167}
&
\textbf{Therapist:} So that could be another adaptive response that they may be enduring a mild stressor, but sometimes they're asleep when you get home.\par \textbf{Patient:} Yeah, that's true.\par \textbf{Therapist:} How bad could it be for them?
&
\emph{Yeah. And they don't know I'm worrying or anxious. They could care less. They don't really know. It's not bothering them.}
&
2 \\
\midrule
\textbf{D}\par{\scriptsize Video 035 Turn 062}
&
\textbf{Therapist:} Yeah. Okay. So driver door seven, passenger door six.\par \textbf{Patient:} Yeah.\par \textbf{Therapist:} All right. How about on the driver's side, starting the car?
&
\emph{That's a nine. That's a nine? I know what happens next.}
&
1 \\
\midrule
\textbf{E}\par{\scriptsize Video 090 Turn 003}
&
\textbf{Therapist:} Okay Kate, we're about 20 minutes into it here. Where are your suds at now?\par \textbf{Patient:} I think I'm down to about a four.\par \textbf{Therapist:} About a four. Yeah. Okay. So let's kind of officially say the Explore the Work is over and I'll take this off now. Yeah. And wow, take a deep breath. Oh, you did it. I did it. Oh my goodness. How do you feel?
&
\emph{I feel proud.}
&
1 \\
\midrule
\textbf{F}\par{\scriptsize Video 087 Turn 057}
&
\textbf{Therapist:} So a long time. So this has been a relatively persistent part of your life, this belief system. So changing it's going to be work. So if you can apply a different philosophy, apply more flexible thinking the next time you're in a meeting, She's still criticizing you in the meetings? Yeah. So if you filtered that through a belief system, now your first urge might be, I would think, to be, I'm being disrespected again, right? It probably comes on that way. You kind of feel it.\par \textbf{Patient:} Yeah, it starts in my stomach and then goes all the way up.\par \textbf{Therapist:} So that would be your first physiological reaction, right, an emotional reaction. But if you can kind of pause that moment and apply more flexible thinking, like I talked about, which is I like respect, but I can survive and function without it. At some level, I can accept that I'm not going to get it all the time. Are there other areas of life where you do get respect?
&
\emph{Yeah. For my friends and my family, I feel like they give me, it's actually mutual respect. They give me respect, I give them respect.}
&
1 \\
\midrule
\textbf{G}\par{\scriptsize Video 002 Turn 014}
&
\textbf{Therapist:} Yeah. So you said that you first noticed the anxiety kicking in about five or six years ago. So you would have been about 19? And what was happening around that time?\par \textbf{Patient:} I guess I'd finished school about a year before that. And I mean, I've always been a bit shy at school, but it was never, ever like it didn't really affect things as much as not at all as much as it has been doing the last few years, I guess. It just started when my friends went to university and I don't know, I felt like I was expected to kind of move out or just do something and then it just kind of escalated from there. I'd just stop going to parties or I'd just say no to invitations or I don't know, I thought my job would help because I mean I love it, I really like doing photography but I'm just kind of getting less and less work as it goes on. Which is my fault, I keep saying no. What kind of work is the most difficult for you in your job? I guess when I have to be working with other people. I don't like that. It's like when I'm trying to... kind of would do an event or something where there's a lot of people there not just photographers but actual you know if I'm like taking photos of people that I don't know I just kind of find myself saying no I mean I want to do them but just it's scary so I just don't do it\par \textbf{Therapist:} So tell me a bit more about what happens then. So let's focus on maybe an invitation to do a job with, you know, it's going to involve photography with a group of people and you're going to have to have social interaction. What kind of thoughts go through your head?
&
\emph{Just, I mean, I get really hot and sweaty and I don't, I feel like everyone's kind of looking at me and thinking that I'm just like really stupid and I feel like they're all going to be staring and thinking I just can't do my job and I'm just an idiot and I mean, they'll think that I just look like a weirdo and I just try and, I mean, I just kind of just try and get out of the situation or just focus on something else, so... That's why I like doing shots where it's just me and the camera because then I can just put all my attention on that instead of having to, you know, be with other people.}
&
4 \\
\end{longtable}
\endgroup
\end{landscape}
\clearpage

\end{document}